\documentclass[runningheads]{llncs}

\usepackage{eccv}

\usepackage{graphicx} 
\usepackage{booktabs}
\usepackage{pifont}
\usepackage{xcolor}
\usepackage[table]{xcolor}

\usepackage{colortbl}
\newcommand{\cellbest}{\cellcolor{red!23}}
\newcommand{\cellsecond}{\cellcolor{orange!26}}

\usepackage{multirow}
\usepackage{tabularx}
\newcolumntype{Y}{>{\centering\arraybackslash}X}
\newcolumntype{C}[1]{>{\centering}p{#1}}
\newcolumntype{Z}{>{\raggedleft\arraybackslash}X}

\usepackage{eccvabbrv}

\usepackage{graphicx}
\usepackage{booktabs}

\usepackage[accsupp]{axessibility}  

\usepackage{hyperref}

\usepackage{orcidlink}

\begin{document}

\title{FLEG: Feed-Forward Language Embedded Gaussian Splatting from Any Views via Compact Semantic Representation} 

\titlerunning{FLEG}

\author{Qijian Tian\inst{1}\orcidlink{0000-0001-7218-4978} \and
Jiayu Ying\inst{2}\orcidlink{0009-0008-3051-7285} \and
Xuhong Wang\inst{3}\orcidlink{0000-0002-0944-8667} \and
Yuan Xie\inst{2}\orcidlink{0000-0001-6945-7437} \and \\
Lizhuang Ma\inst{1,2}\orcidlink{0000-0003-1653-4341} \and 
Xin Tan\inst{2,3}\orcidlink{0000-0001-9346-1196}}

\authorrunning{Q.~Tian et al.}

\institute{Shanghai Jiao Tong University, Shanghai, China \\
\email{tianqijian@sjtu.edu.cn, ma-lz@cs.sjtu.edu.cn} \and 
East China Normal University, Shanghai, China \\
\email{yingjiayu12@gmail.com, xieyuan8589@foxmail.com, xtan@cs.ecnu.edu.cn} \and
Shanghai Artificial Intelligence Laboratory, Shanghai, China \\
\email{wangxuhong@pjlab.org.cn} \\
[2.0em]
\href{https://fangzhou2000.github.io/projects/fleg}{
    \textcolor{magenta}{
        \textbf{\nolinkurl{https://fangzhou2000.github.io/projects/fleg}}
    }
}}

\maketitle

\begin{abstract}

We present FLEG, a feed-forward network that reconstructs language-embedded 3D Gaussians from multi-view images.
Previous feed-forward language-embedded Gaussian reconstruction methods are restricted to a fixed number of input views and typically attach a language-aligned semantic embedding to each Gaussian, resulting in impractical input settings and semantic redundancy. In contrast, we introduce a geometry-semantic dual-branch distillation framework that supports a variable number of uncalibrated and unposed multi-view images as input. We also propose a novel-view-based distillation strategy during training that mitigates overfitting to input views.
In addition, we observe that semantic representations are significantly sparser than geometric ones, and per-Gaussian language embedding is unnecessary. To exploit this sparsity, we design a decoupled language embedding strategy that represents language information with a sparse set of semantic Gaussians, rather than attaching embeddings to every Gaussian. Compared with dense pixel-aligned per-Gaussian embedding schemes, our method uses only 5\% of the language embeddings while maintaining comparable semantic fidelity, effectively reducing storage costs.
Extensive experiments demonstrate that FLEG outperforms state-of-the-art feed-forward reconstruction and language-embedded Gaussian methods in both reconstruction quality and language-aligned semantic representation.\par
\end{abstract}    
\section{Introduction}
\label{sec:intro}

\begin{figure*}[t!]
    \centering
    \includegraphics[width=1.0\textwidth]{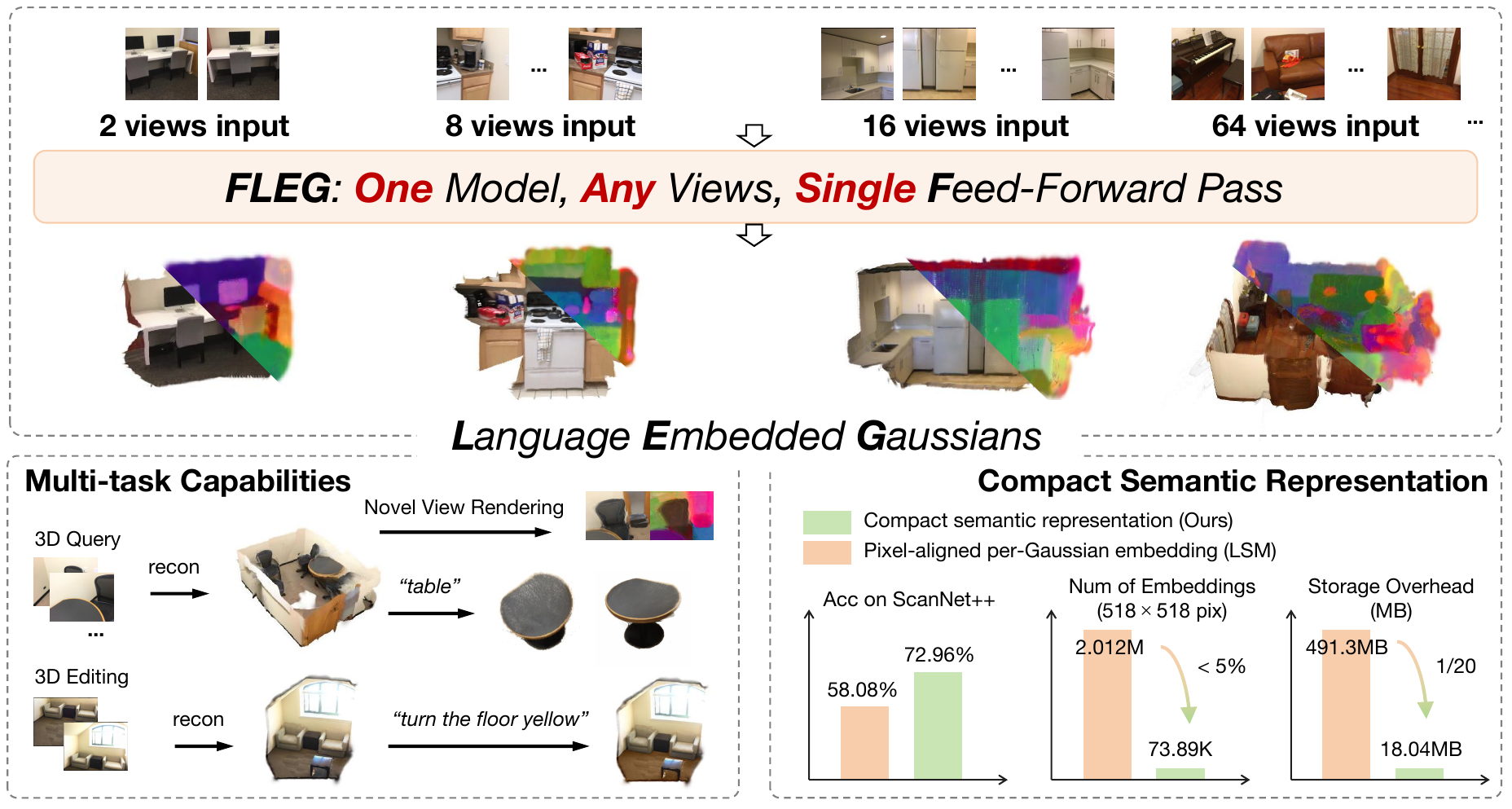}
    \caption{FLEG reconstructs language-embedded Gaussians in a single feed-forward pass from a variable number of uncalibrated and unposed images, supporting both sparse and dense views in one model with the compact semantic representation that avoids per-Gaussian language embedding and significantly reduces storage overhead.}
    \label{fig: teaser}
\end{figure*}

Embedding language-aligned semantic features into 3D representations is essential for interaction with 3D environments in a human-like manner, which enables query, editing, and other interactions with 3D through natural language. It has various applications, such as robotic navigation~\cite{vision-and-language_navigation, robot_navigation}, manipulation~\cite{language-guided_manipulation, open-world_object_manipulation}, and augmented/virtual reality~\cite{augmented_reality}. Constructing such 3D language fields requires lifting 2D observations into coherent 3D geometry and appearance while jointly embedding semantics to achieve open-vocabulary interaction, which involves both 3D reconstruction and vision-language modeling.\par

With the rapid development of 3D Gaussian Splatting~\cite{3dgs, 3dgs_survey} (3DGS), many works~\cite{langsplat, feature3dgs, opengaussian, fastlgs, langscenex} embed semantics from 2D vision-language models into Gaussians and build language fields through per-scene optimization. Although the differentiable rasterization of 3DGS enables fast rendering, the per-scene optimization remains inefficient for language field reconstruction, as the optimization often takes minutes or even hours, making these methods impractical for real-time applications such as robotic navigation and manipulation. Furthermore, these methods generally require dense multi-view observations and struggle in sparse-view settings, which are more prevalent in practical deployments. \par

In contrast to per-scene optimization, recent feed-forward reconstruction methods~\cite{dust3r, mast3r, vggt} achieve approximately real-time 3D reconstruction from multi-view images to 3D point clouds, providing a promising direction toward more practical language-embedded Gaussian reconstruction. Inspired by this paradigm, several works attempt feed-forward language-embedded Gaussian reconstruction. However, these methods are typically restricted to only two input views~\cite{lsm, uniforward, gsemsplat, semanticsplat, spatialsplat} or require additional camera parameters~\cite{uni3r}, which limits their practical applicability in real-world scenarios. Achieving feed-forward semantic reconstruction from flexible multi-view inputs without camera parameters remains challenging, since scalable geometric and semantic supervision (e.g., 3D scans, dense depth, and corresponding semantic labels) is prohibitively costly, limiting training scale and data diversity. Moreover, another advantage of feed-forward methods is the capability to reconstruct large-scale assets rapidly, while existing methods typically predict pixel-aligned Gaussians and attach a distinct language embedding to every Gaussian. The number of required language embeddings scales linearly with the number of input views, leading to semantic redundancy and substantial storage overhead for assets. \par

In this paper, we propose FLEG, a feed-forward network that reconstructs language-embedded Gaussians from uncalibrated and unposed multi-view images, as shown in \cref{fig: teaser}.
To address the scarcity of 3D data annotated with both geometric and semantic labels, we leverage VGGT~\cite{vggt}, a 3D foundation model, to provide geometric supervision and distill semantics from CLIP features~\cite{clip}. Although several recent methods~\cite{anysplat, uni3r} also leverage geometric supervision from VGGT, they restrict supervision to the observed input views. In contrast, we propose a novel-view-based distillation that selects reliable novel views for both geometric and rendering supervision, effectively preventing overfitting to input views and providing novel-view supervision that enforces multi-view reconstruction consistency.
For semantic distillation, we align rendered embeddings with CLIP features and introduce instance-guided contrastive learning. Embeddings from the same instance are pulled toward the instance’s average CLIP feature, while embeddings from different instances are pushed apart, improving both language alignment and cross-instance discriminability. The geometric and semantic distillation form a dual-branch distillation framework, allowing our network to be trained without any 3D ground truth or semantic labels. The 2D instance masks required by contrastive learning can be easily obtained using off-the-shelf segmentation models~\cite{sam2}. This design enables training on arbitrary video sequences and multi-view images, highlighting the potential for large-scale training. 
In addition, we observe that semantic representations are significantly sparser than geometric ones. This sparsity suggests that the prevailing paradigm of attaching an embedding to every Gaussian is unnecessary, leading to substantial semantic redundancy and storage overhead. To address this, we introduce the Decoupled Gaussian Language Embedding (DGLE) module, which parametrically fuses language embeddings into a much sparser set of semantic Gaussians, yielding a separate compact semantic representation. Experiments demonstrate that, compared to existing methods that attach embeddings to each pixel-aligned Gaussian, our method achieves comparable semantic fidelity using less than 5\% of the embedding number. This confirms the sparsity of semantic representations and substantially reduces the storage overhead for reconstructed assets. \par

Through comprehensive design across the training framework, learning strategy, and semantic representation, our FLEG simultaneously achieves accurate geometric reconstruction, high-fidelity rendering, and expressive language-aligned semantics with substantially fewer embeddings, supporting both dense and sparse input views. In summary, our key contributions are as follows:
\begin{itemize}
    \item \textbf{Dual-Branch Distillation Framework.} We train the feed-forward language-embedded Gaussian reconstruction network without any 3D ground-truth or semantic labels, supporting arbitrary input views without camera parameters. In the geometric branch, we propose a novel-view-based distillation that prevents overfitting to observed views. In the semantic branch, we introduce instance-guided contrastive learning together with CLIP feature alignment to enhance embedding discriminability across instances. \par
    
    \item \textbf{Decoupled Gaussian Language Embedding (DGLE) module.} We propose DGLE that parametrically fuses semantic embeddings into a much sparser set of separate compact semantic Gaussians. Compared to the prevailing paradigm of attaching embeddings to each pixel-aligned Gaussian, it achieves comparable semantic fidelity using less than 5\% of the embeddings, significantly reducing storage overhead. \par
    
    \item \textbf{Multi-task capabilities.} Our FLEG simultaneously supports multiple tasks through language-embedded Gaussians, including novel-view synthesis, open-vocabulary object query, and 3D editing. Experiments show that FLEG achieves the best overall performance against both feed-forward Gaussian reconstruction and language-embedded Gaussian reconstruction methods. \par
\end{itemize}

\section{Related Work}

\subsection{Feed-Forward Reconstruction}

Typical per-scene optimization reconstruction methods like NeRF~\cite{nerf} and 3DGS~\cite{3dgs} achieve high-quality rendering while often taking minutes or even hours to optimize a single scene, making them unsuitable for real-time applications. Recent advances in feed-forward reconstruction have substantially improved the reconstruction efficiency, both for Gaussian-based methods~\cite{pixelsplat, mvsplat, drivingforward, splatt3r, noposplat, pf3plat, spfsplat, selfsplat, flare, anysplat} and point-map-based methods~\cite{dust3r, mast3r, cut3r, vggt, da3}. However, these methods mainly focus on geometry and appearance reconstruction and do not include semantics, limiting natural language interaction with the 3D scene. Another line of work explores scene understanding with pretrained feed-forward 3D backbones~\cite{pointtransformer, swin3d}, but these methods require explicit 3D inputs and focus on 3D perception rather than reconstruction. In contrast, our feed-forward method recovers high-quality geometry and appearance from arbitrary input views while simultaneously embedding semantics into the 3D representation, achieving joint 3D reconstruction and semantic understanding. \par

\subsection{Language Embedded Gaussian Splatting}

Vision-language models provide effective 2D semantic priors for language-aligned scene understanding~\cite{clip, lseg, clipsp}. With recent advances in 3DGS, several methods~\cite{langsplat, legaussians, opengaussian} have explored leveraging these vision-language models to embed language-aligned features into Gaussians and construct language embedded fields. Subsequent works~\cite{fastlgs, langsplatv2} improve the query efficiency and achieve higher rendering speed. DF-3DGS~\cite{df-3dgs} further explores the sparsity of semantic features for 3DGS. However, these per-scene optimization methods remain computationally expensive for reconstruction, limiting their practical applicability in real-world scenarios and the capability to produce large-scale reconstruction assets. Following recent advances in feed-forward reconstruction, several methods~\cite{lsm, uniforward, gsemsplat, semanticsplat, spatialsplat} attempted to achieve feed-forward language-embedded field construction. However, these methods can only process two input views at once and struggle to handle dense view inputs. Uni3R~\cite{uni3r} extends to multi-view settings but remains view-specific, which is less flexible and requires additional camera parameters. Our method overcomes these limitations, enabling feed-forward language field reconstruction from arbitrary uncalibrated and unposed multi-view images. A recent work, IGGT~\cite{iggt}, performs feed-forward instance-level point map reconstruction from multi-view images. It predicts pixel-wise point maps with instance features, and the point map representation does not support novel-view synthesis. In contrast, FLEG predicts Gaussians with sparse language embeddings, enabling high-fidelity novel-view synthesis. \par

\begin{figure*}[t!]
    \centering
    \includegraphics[width=1.0\textwidth]{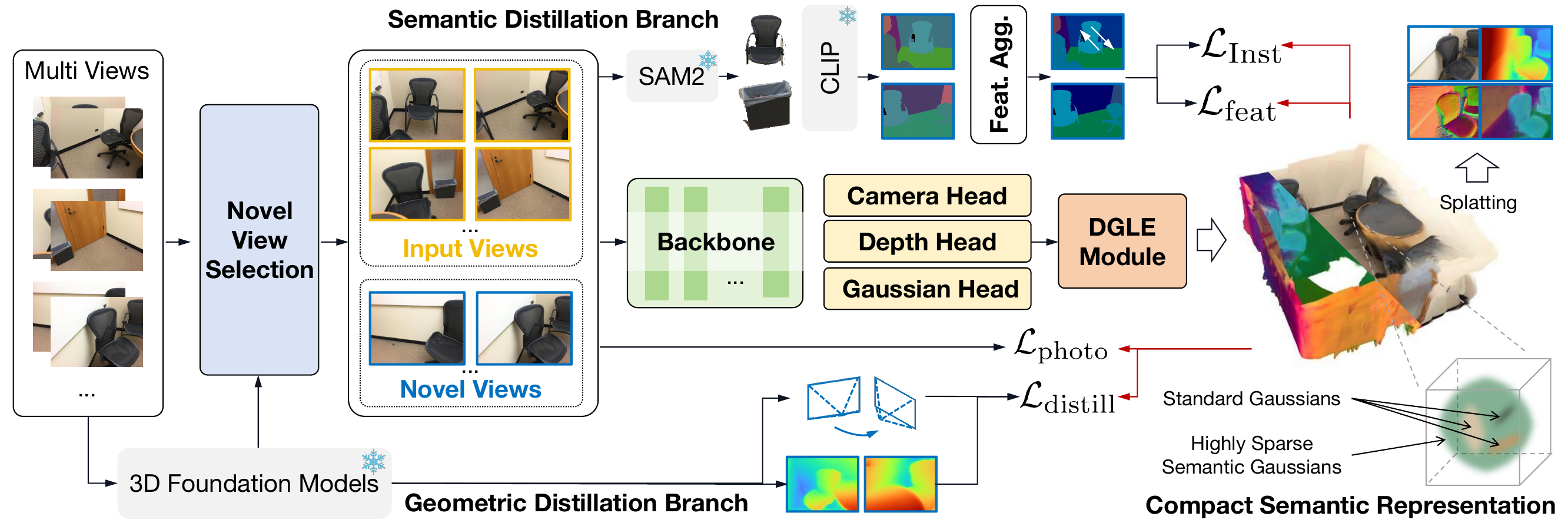}
    \caption{Overview of FLEG. We propose a dual-branch distillation to train the network without 3D ground truth and semantic labels. We also propose a Decoupled Gaussian Language Embedding (DGLE) module that produces the compact semantic representation, which significantly reduces semantic redundancy and storage overhead.}
    \label{fig: main}
\end{figure*}

\section{Method}
We propose FLEG, a feed-forward network that reconstructs language-embedded Gaussians from arbitrary uncalibrated and unposed multi-view images. It simultaneously recovers accurate geometry, high-fidelity appearance, and language-aligned semantics, supporting both sparse (as few as 2) and dense (more than 30) input views.
Our proposed dual-branch distillation framework enables training without costly 3D ground-truth geometry or semantic annotations. The Decoupled Gaussian Language Embedding (DGLE) module avoids attaching embeddings to each Gaussian, significantly reducing storage overhead.
The resulting language-embedded Gaussians support multiple downstream tasks, including novel-view synthesis, open-vocabulary object queries, and 3D editing. An overview of FLEG is shown in \cref{fig: main}.

\subsection{Network Architecture}
Our FLEG employs a large transformer following VGGT~\cite{vggt}, which first patchifies each image into a set of tokens using DINOv2~\cite{dinov2}. A learnable camera token and four register tokens are appended to each view’s token sequence. The combined tokens are processed through 24 layers of global and frame-wise attention. We adopt VGGT’s depth and camera heads to predict depth maps and camera parameters. The predicted depths are then back-projected into 3D points $\mu$ in the world coordinate system using the predicted camera parameters. Subsequently, we design a DPT~\cite{dpt}-based Gaussian head to predict additional attributes of the language-embedded Gaussians, including scale $s \in \mathbb{R}^3$, rotation quaternion $r \in \mathbb{R}^4$, opacity $\sigma \in \mathbb{R}^{+}$, spherical harmonic (SH) coefficients $c \in \mathbb{R}^{3 \times (k+1)^2}$ for representing color with degree-$k$ SH, and a $D$-dimensional language-aligned semantic feature $emb \in \mathbb{R}^D$. A voxelization module aggregates these attributes into a 3D voxel-based representation, forming the standard Gaussians $G = \{\mu, s, r, \sigma, c\}$. Finally, our DGLE module parametrically fuses language embeddings into a much sparser set of semantic Gaussians, resulting in $G_{sem} = \{\mu, s, r, \sigma, emb\}$, effectively decoupling semantic information from the dense geometric Gaussians. \par

\subsection{Dual-Branch Distillation Framework}

To train the network without costly 3D ground-truth and semantic annotations, we propose a geometric-semantic dual-branch distillation framework. In the geometric distillation branch, we leverage a frozen VGGT~\cite{vggt} and use its geometric predictions to both select reliable novel views and provide pseudo-label supervision for the rendered novel views, while also applying the photometric reconstruction loss. In the semantic distillation branch, we first obtain 2D instance masks using SAM2~\cite{sam2} and then use CLIP~\cite{clip} to extract language-aligned instance features for distillation. In addition to direct CLIP feature alignment, we introduce an instance-guided contrastive learning algorithm that distinguishes embeddings of different instances while preserving language alignment, improving instance-level semantic discriminability. This dual-branch framework enables training on arbitrary video sequences and multi-view image collections, underscoring the potential for large-scale training. \par

\subsubsection{Novel-View-Based Distillation}

Although several recent works~\cite{anysplat, uni3r} also leverage VGGT for geometric distillation, they constrain supervision to input views only, which may induce overfitting to those views and degrade cross-view consistency. Existing feed-forward Gaussian reconstruction methods typically adopt novel-view supervision to mitigate this problem. However, such methods either depend on ground-truth camera poses~\cite{pixelsplat, mvsplat, splatt3r, noposplat} or employ complex self-supervised training~\cite{pf3plat, selfsplat, spfsplat} that often yields unstable optimization and makes training for arbitrary-view inputs more challenging. To address these limitations, we propose the novel-view-based distillation to combine VGGT distillation with novel-view supervision, enabling novel-view supervision without access to camera parameters while preserving training stability. \par

Specifically, given $N$ input images $\{I_i\}_{i=1}^{N}$, we obtain the predicted 3D point maps $\{P_i\}_{i=1}^{N}$ and camera parameters $\{K_i\}_{i=1}^{N}$, $\{E_i\}_{i=1}^{N}$ from VGGT. During training, we randomly sample $N_c$ images from $\{I_i\}_{i=1}^{N}$ as context views $\{I_i\}_{i=1}^{N_c}$ for network input, where $2 \le N_c \le N$. For target views used for novel-view supervision, we select images from $\{I_i\}_{i=1}^{N}$ that are sufficiently covered by context views to ensure that novel views remain observable and do not introduce many unseen regions that may degrade supervision quality. We project point maps of context views $\{P_i\}_{i=1}^{N_c}$ onto $\{I_i\}_{i=1}^{N}$ to obtain the binary coverage masks $\{M_i\}_{i=1}^{N}$:
\begin{equation}
    \{M_i\}_{i=1}^{N} = \Pi(\{P_i\}_{i=1}^{N_c}, \{K_i\}_{i=1}^{N}, \{E_i\}_{i=1}^{N}),
\end{equation}
where $\Pi$ is the projection function that maps 3D points from world coordinates to camera coordinates, and $\{M_i\}_{i=1}^{N}$ represent the binary masks of $\{I_i\}_{i=1}^{N}$. \par

Target views are selected as the images with aggregated coverage masks exceeding a predefined threshold $\tau$:
\begin{equation}
    \{I_i\}_{i=1}^{N_t} = \{I_j | \operatorname{cov}(M_j) > \tau\},
\end{equation}
where $\operatorname{cov}(\cdot)$ is the coverage ratio for each image:
\begin{equation}
    \operatorname{cov}(M_j) = \frac{1}{H \times W} \sum_{h=1}^{H} \sum_{w=1}^{W} \mathbb{I}\big(M_j(h, w) > 0\big).
\end{equation}
With the selected target views, we render the novel view image using $\{E_i\}_{i=1}^{N_t}$, and apply photometric supervision with an additional LPIPS loss~\cite{lpips} on novel views to enhance consistency across views:
\begin{equation}
\begin{split}
    \mathcal{L}_{\mathrm{photo}} = \eta_1 \frac{1 - \operatorname{SSIM}(I, \hat{I})}{2} + (1 - \eta_1) \lVert I - \hat{I}\rVert + \eta_2\mathcal{L}_{\mathrm{LPIPS}}, \quad I \in \{I_i\}_{i=1}^{N_t}. 
\end{split}
\end{equation}
Note that $\{E_i\}_{i=1}^{N_t}$ is only used for training, and the model is fully pose-free during inference. \par

\subsubsection{Semantic Distillation with Instance-guided Contrastive Learning}

The geometric distillation branch enables the network to learn geometry and appearance from $\mathcal{L}_{\mathrm{distill}}$, which includes an MSE loss for depth maps and a Huber loss for camera parameters, and $\mathcal{L}_{\mathrm{photo}}$, respectively. To further learn language-aligned semantics during training and predict language-embedded Gaussians, we lift the CLIP features~\cite{clip} from 2D to 3D through feature distillation, also without relying on any 3D annotations. \par

We first employ SAM2~\cite{sam2} to obtain instance masks of video sequences, then we extract language-aligned features for each instance using CLIP~\cite{clip}. The CLIP feature for each instance is applied to the pixels within its corresponding instance mask, generating the pixel-wise language-aligned feature as the target feature $F \in \mathbb{R}^{N_t \times C \times H \times W}$. The predicted language-embedded Gaussians with D-dimensional embeddings $emb$ are rendered into 2D feature maps $\hat{F} \in \mathbb{R}^{N_t \times C \times H \times W}$, which are passed through a lightweight MLP, and we compute cosine similarity loss with the corresponding target features: 
\begin{equation}
\begin{split}
    \mathcal{L}_{\mathrm{feat}} =  1 - \frac{\hat{F} \cdot F}{\lVert\hat{F}\rVert \lVert F \rVert},
\end{split}
\end{equation}
In addition, to address the inconsistency of instance masks across multi-views, we propose a 3D feature aggregation module to obtain more consistent multi-view target features. Specifically, we first use the 3D point maps predicted by VGGT to lift the 2D pixel-wise language-aligned features into 3D space. Based on the assumption that pixels mapped to the same spatial location should share similar features, we voxelize the 3D space and average the features within each voxel, which are subsequently projected back to the 2D image, and we further average the features within each instance mask. This process enhances the multi-view consistency of CLIP-extracted features. \par

However, we observe that purely applying CLIP feature alignment lacks instance-level discrimination, resulting in blurred boundaries between different instances. To address this problem and better align the 2D semantics with 3D representations, we introduce an instance-guided contrastive learning algorithm. Specifically, given a feature map $F \in \mathbb{R}^{N_t \times C \times H \times W}$ and corresponding instance mask $M \in \mathbb{R}^{N_t \times H \times W}$, 
we first compute the mean feature vector for each instance as the anchor feature $\mathbf{f}^{\text{ins}}_k \in \mathbb{R}^{C}$. For each sampled pixel, the rendered embedding vector $\hat{\mathbf{f}^{\text{pix}}_j} \in \mathbb{R}^{C}$ is regarded as a positive sample if it belongs to the same instance as the anchor, and as a negative sample otherwise. The instance-guided contrastive loss is formulated as:
\begin{equation}
\mathcal{L}_{\mathrm{inst}}
= -\frac{1}{K} \sum_{k=1}^{K} 
\log 
\frac{
\sum_{j \in \mathcal{P}_k} 
\exp\left(\operatorname{sim}\left(\mathbf{f}^{\mathrm{ins}}_k, \hat{\mathbf{f}^{\mathrm{pix}}_j}\right)/\alpha\right)
}{
\sum_{j} 
\exp\left(\operatorname{sim}\left(\mathbf{f}^{\mathrm{ins}}_k, \hat{\mathbf{f}^{\mathrm{pix}}_j}\right)/\alpha\right)
},
\end{equation}
where $\mathcal{P}_k$ is the set of pixels belonging to the same instance as the $k$-th anchor feature, and $\alpha$ is the temperature parameter. \par

This design promotes the 2D-3D feature alignment and enhances instance-level semantic discrimination. In addition, it substantially reduces computational cost compared with general pixel-wise contrastive learning that requires pairwise similarity computations among all $N_p$ pixels, leading to a complexity of $\mathcal{O}(N_p^2)$. In contrast, our algorithm aggregates pixels into $K$ instance-level anchors and computes similarities only between these anchors and $N_p$ pixels, resulting in a complexity of $\mathcal{O}(N_pK)$, where typically $K \ll N_p$. With additional pixel sampling, the cost can be reduced further. This substantially reduces computation while preserving strong instance-level discrimination, making it well-suited for feed-forward training. \par

\subsection{Decoupled Gaussian Language Embedding}

Typical feed-forward reconstruction methods often produce pixel-wise predictions. Although dense predictions are inherently redundant, the redundancy is relatively minor for geometry and appearance modeling. However, semantic information is intrinsically much sparser than geometric information, representing semantics at the same density as geometry introduces substantial redundancy and considerable storage overhead. This mismatch in representation density conflicts with prevailing language-embedded Gaussian methods that associate a distinct language embedding with each Gaussian. The problem is particularly severe for feed-forward reconstruction. Under pixel-aligned Gaussian prediction, the number of required language embeddings grows linearly with the number of input views, resulting in pronounced semantic redundancy and substantial storage overhead for reconstructed assets. \par

To this end, we propose a decoupled Gaussian language embedding and define a novel compact semantic representation to address the mismatch in representation density between geometric and semantic information. Specifically, given the attributes predicted by the Gaussian head, we first construct pixel-wise standard Gaussians as initial Gaussians $G_{\mathrm{pix}}$ that represent geometry and appearance. Following AnySplat~\cite{anysplat}, we apply a voxelization module to reduce the overlapping geometric Gaussians in the 3D coordinate space:
\begin{equation} \label{eq: voxel}
    V_i = \left\lfloor \frac{\mu_i}{v_{\mathrm{geo}}} \right\rceil,
    \quad i=1,\ldots,N_c \times H \times W,
\end{equation}
where $V_i$ is the voxel index of the $i$-th initial Gaussian, $v_{\mathrm{geo}}$ is the voxel size, and $N_{\mathrm{geo}}$ denotes the number of occupied voxels after voxelization. The attributes of initial Gaussians and pixel-wise language embeddings are averaged within corresponding voxels using softmax-normalized confidence weights:
\begin{equation} \label{eq: average}
    \bar{x}_n = \sum_{i:V_{i}=n} \frac{\exp(conf_i)}{\sum_{j:V_{j}=n}\exp(conf_j)} x_{i},
\end{equation}
where $x_i \in \{\mu_i, s_i, r_i, \sigma_i, c_i, conf_i, emb_i\}_{i=1}^{N_c \times H \times W}$ and $conf_i$ is the additional predicted Gaussian confidence from the Gaussian head. We define the voxelized standard Gaussians as $G_{\mathrm{voxel}}$.

Although voxelization reduces redundancy for geometry and appearance, the density remains redundant for semantic representation. To exploit the sparsity of semantics, we introduce a compact representation, named semantic Gaussians, which is decoupled from the voxelized standard Gaussians $G_{\mathrm{voxel}}$:
\begin{equation}
    G_{\mathrm{sem}} = \{\mu, s, r, \sigma, emb\}.
\end{equation}
Compared with standard Gaussians, the semantic Gaussians omit the color attributes encoded by spherical harmonics and instead incorporate language embeddings, since their purpose is to model semantics rather than appearance. \par

The parameters of semantic Gaussians are parametrically derived from the attributes of $G_{\mathrm{voxel}}$ and voxelized language embeddings. Specifically, we first define a much larger voxel with voxel size $v_{\mathrm{sem}}$ according to \cref{eq: voxel}. $\mu_i$, $\sigma_i$ and $emb_i$ of $G_{\mathrm{voxel}}$ are aggregated through the same softmax average in \cref{eq: average} to derive $\mu_j$, $\sigma_j$ and $emb_j$ of $G_{\mathrm{sem}}$. The average weight is denoted as $w_i$.

For the scale $s_j$ and rotation $r_j$ that determine the shape of each semantic Gaussian $G_{\mathrm{sem}}$, we do not apply the same averaging used for the aggregation of position, opacity, and embedding. Since semantic Gaussians are constructed with higher sparsity, they are encouraged to spatially cover the corresponding standard Gaussians to maintain semantic consistency across regions. Moreover, as semantic information exhibits weaker anisotropic variation compared to geometric appearance, we employ isotropic semantic Gaussians to achieve a compact and stable representation. \par

For all standard Gaussians in $G_{\mathrm{voxel}}$ assigned to the $j$-th semantic voxel, their covariance matrices $\Sigma_i = R_i \mathrm{diag}(s_i^2) R_i^\top$ are fused using a moment-matching scheme with softmax-normalized confidence weights $w_i$:
\begin{equation}
    \Sigma_j = \sum_{i:\, V_i^{\mathrm{sem}}=j} w_i \left( \Sigma_i + (\mu_i - \mu_j)(\mu_i - \mu_j)^{\top} \right),
\end{equation}
where $V_i^{\mathrm{sem}}$ denotes the semantic voxel index of the $i$-th standard Gaussian. Instead of retaining the full anisotropic covariance, we approximate the fused covariance as isotropic using its trace:
\begin{equation}
    s_j = \sqrt{\frac{\operatorname{Tr}(\Sigma_j)}{3}} \cdot \mathbf{1}, 
    \qquad r_j = [1, 0, 0, 0],
\end{equation}
where $s_j$ and $r_j$ denote the scale and rotation of the semantic Gaussian, respectively. 
This isotropic approximation ensures that each semantic Gaussian effectively spans the corresponding geometric region while avoiding unnecessary anisotropic computation. \par

Experiments show that our compact semantic representation achieves comparable or even superior semantic performance with only 5\% of the number of per-pixel embeddings, demonstrating that the decoupled Gaussian language embedding effectively reduces semantic redundancy and storage overhead. \par

\subsection{Training Objectives}

The proposed dual-branch distillation framework enables training our model on arbitrary video data or multi-view images without requiring camera parameters, depth maps, 3D ground truth, or costly semantic annotations, making scalable model training possible. Overall, we utilize the following loss function:
\begin{equation}
\mathcal{L} = 
\lambda_1\mathcal{L}_{\mathrm{photo}} + 
\underbrace{\lambda_2 \mathcal{L}_{\mathrm{distill}}}_{\text{geo. distillation}} +
\underbrace{\lambda_3 \mathcal{L}_{\mathrm{feat}} + \lambda_4 \mathcal{L}_{\mathrm{inst}}}_{\text{semantic distillation}},
\end{equation}
where $\mathcal{L}_{\mathrm{photo}}$ provides the basic photometric loss, and $\mathcal{L}_{\mathrm{distill}}$ belongs to the geometric branch of the proposed dual-branch distillation framework, while $\mathcal{L}_{\mathrm{feat}}$ and $\mathcal{L}_{\mathrm{inst}}$ belong to the semantic branch, which improve the quality of geometry and semantics, respectively. \par 
\section{Experiments}

\subsection{Implementation Details}
We initialize the transformer layers, camera head, and depth head with weights from pretrained VGGT~\cite{vggt} and randomly initialize the Gaussian head. The whole network contains approximately 1B parameters. During training, inputs are resized to a maximum resolution of 518 pixels on the longer side with aspect ratios sampled between 0.5 and 1.0 and augmented via random horizontal flipping. For each iteration, we set $N = 12$ and randomly select $2 \le N_c \le N$ as the input views. We train our network on ScanNet~\cite{scannet} and ScanNet++~\cite{scannet++}, which is consistent with the training data of our main comparison methods, LSM~\cite{lsm} and Uni3R~\cite{uni3r}. Our network is trained on 8 NVIDIA H200 GPUs for approximately two days, which is also comparable to the training resources required by the main baselines. In the training objectives, $\mathcal{L}_{distill}$ includes the MSE loss between the rendered depth and the pseudo-label depth from VGGT and the Huber loss between the predicted camera parameters and their pseudo-labels from VGGT, with the weights set to 0.1 and 10.0, respectively. The hyperparameters in overall training loss are set as $\eta_1=0.5, \eta_2=0.2, \lambda_1=2.0, \lambda_2=1.0, \lambda_3=0.5, \lambda_4=0.05$. \par

\subsection{Evaluation Details} 
To jointly evaluate the quality of reconstruction and semantic embeddings, we evaluate reconstruction quality via novel view synthesis and introduce open-vocabulary semantic segmentation under novel views to evaluate the semantic alignment with language. For reconstruction, we use standard reconstruction metrics, including PSNR, SSIM, and LPIPS. For open-vocabulary segmentation, we use mean Intersection-over-Union (mIoU) and mean Location Accuracy (Acc.) following LangSplat~\cite{langsplat}. We conduct the evaluation on the validation sets of the ScanNet~\cite{scannet} and ScanNet++\cite{scannet++} datasets. We uniformly sample frames to form the input views, with the number of input views varying from 2 to 16. Novel views for evaluation are selected from the remaining frames, with their number set to half of the input views. For pose-free evaluation, we adopt the same alignment strategy from AnySplat~\cite{anysplat} to our method. \par

\subsection{Baselines}
We compare our method with existing state-of-the-art feed-forward Gaussian reconstruction methods (AnySplat~\cite{anysplat}), feed-forward language-embedded Gaussian reconstruction methods (LSM~\cite{lsm} and Uni3R~\cite{uni3r}), and per-scene optimized language-embedded Gaussian reconstruction methods (LangSplat~\cite{langsplat} and LangScene-X~\cite{langscenex}). We also compare with the 2D semantic models (LSeg~\cite{lseg} and CLIP~\cite{clip}) commonly used for distillation in language-embedded Gaussian reconstruction methods. All baselines are open-source. We adopt their officially released implementations and pretrained models, strictly following the provided evaluation protocols to ensure fair comparison. Some baselines support only a fixed number of input views, such as LSM and Uni3R, whose officially available pretrained models handle only two views. Extending these models to more views is non-trivial, as it would require architectural modifications and retraining. To ensure fair comparison across varying input-view settings, we use their official two-view pretrained models, selecting the two nearest input views to the target view as input. This evaluation protocol has also been adopted in prior work~\cite{flare}. We find that it performs better than optimization-based post-processing and remains faithful to the original baseline configurations. \par

\begin{table}[t!]
\centering
\caption{Comparison of feed-forward methods on ScanNet++ and ScanNet for novel view synthesis and open-vocabulary query segmentation across different input views. The \colorbox{red!23}{best} and \colorbox{orange!26}{runner-up} results are highlighted in red and orange.}
\scriptsize
\begin{tabularx}{0.99\textwidth}{l *{5}{Y} *{5}{Y}}
\toprule

& \multicolumn{5}{c}{ScanNet++~\cite{scannet++}} & \multicolumn{5}{c}{ScanNet~\cite{scannet}} \\
\cmidrule(r){2-6} \cmidrule(l){7-11}

& mIoU$\uparrow$ & Acc.$\uparrow$ & PSNR$\uparrow$ & SSIM$\uparrow$ & LPIPS$\downarrow$
& mIoU$\uparrow$ & Acc.$\uparrow$ & PSNR$\uparrow$ & SSIM$\uparrow$ & LPIPS$\downarrow$ \\
\cmidrule(r){1-6} \cmidrule(l){7-11}

\multicolumn{1}{c}{} & \multicolumn{5}{c}{2 views} & \multicolumn{5}{c}{2 views} \\
\cmidrule(r){1-6} \cmidrule(l){7-11}
~LSeg\cite{lseg}             & \cellsecond 45.28 & 59.21             & -                  & -                  & -                  & \cellsecond 42.04 & \cellsecond 60.49 & -                  & -                  & -                 \\ 
\cite{sam2}+CLIP~\cite{clip} & 39.58             & \cellsecond 62.39 & -                  & -                  & -                  & 34.67             & 58.52             & -                  & -                  & -                 \\ 
~AnySplat\cite{anysplat}      & -                 & -                 & \cellsecond 26.81  & \cellsecond 0.8532 & \cellsecond 0.1833 & -                 & -                 & 23.03              & 0.7701             & \cellsecond0.2924            \\ 
~LSM\cite{lsm}                & 39.36             & 58.35             & 19.87              & 0.7359             &           0.4120   & 39.22             & 59.05             & 22.07              & 0.7606             & 0.3912  \\ 
~Uni3R\cite{uni3r}            & 39.54             & 58.59             & 24.95              & 0.7956             & 0.2837             & 39.24             & 58.79             & \cellsecond 23.92 & \cellsecond 0.7737  &  0.3212 \\ 
~FLEG (Ours)                  & \cellbest 46.61   & \cellbest 71.18   & \cellbest 28.14    & \cellbest 0.8718   & \cellbest 0.1451   & \cellbest 48.01   & \cellbest 73.59   & \cellbest 25.05    & \cellbest 0.8076   & \cellbest 0.2715  \\
\cmidrule(r){1-6} \cmidrule(l){7-11}

\multicolumn{1}{c}{} & \multicolumn{5}{c}{4 views} & \multicolumn{5}{c}{4 views} \\
\cmidrule(r){1-6} \cmidrule(l){7-11}
~LSeg~\cite{lseg}            & \cellsecond 45.57 & \cellsecond 58.64 & -                  & -                  & -                  & \cellsecond 41.44 & \cellsecond 60.90 & -                  & -                  & -                 \\
\cite{sam2}+CLIP~\cite{clip} & 37.60             & 57.66             & -                  & -                  & -                  & 32.72             & 56.45             & -                  & -                  & -                 \\
~AnySplat~\cite{anysplat}    & -                 & -                 & \cellsecond 25.28  & \cellsecond 0.8320 & \cellsecond 0.2050 & -                 & -                 & \cellsecond 22.56  & \cellsecond 0.7735 & \cellsecond 0.3191 \\
~LSM~\cite{lsm}              & 41.33             & 57.91             & 19.64              & 0.7359             & 0.4144             & 38.99             & 59.46             & 21.56              & 0.7626             & 0.4041  \\
~Uni3R~\cite{uni3r}          & 41.87             & 58.28             & 23.22              & 0.7816             & 0.3080             & 37.41             & 58.16             & 19.63              & 0.6909             & 0.4212            \\
~FLEG (Ours)                 & \cellbest 48.09   & \cellbest 74.46   & \cellbest 27.50    & \cellbest 0.8627   & \cellbest 0.1557   & \cellbest 48.85   & \cellbest 74.20   & \cellbest 24.70    & \cellbest 0.8123   & \cellbest 0.2933 \\
\cmidrule(r){1-6} \cmidrule(l){7-11}

\multicolumn{1}{c}{} & \multicolumn{5}{c}{8 views} & \multicolumn{5}{c}{8 views} \\
\cmidrule(r){1-6} \cmidrule(l){7-11}
~LSeg~\cite{lseg}             & \cellsecond 44.26 & 58.29             & -                 & -                  & -                  & \cellsecond 41.18 & \cellsecond 59.46  & - & - & -     \\
\cite{sam2}+CLIP~\cite{clip} & 37.84             & \cellsecond 59.92 & -                 & -                  & -                  & 32.71             & 55.16              & - & - & -     \\
~AnySplat~\cite{anysplat}     & -                 & -                 & \cellsecond 24.82 & \cellsecond 0.8147 & \cellsecond 0.2128 & -                 & -                  & \cellsecond 22.04 & \cellsecond 0.7622 & \cellsecond 0.3471 \\
~LSM~\cite{lsm}               & 40.16             & 58.88             & 19.28             & 0.7335             & 0.4144             & 38.61             & 58.50              & 20.82             &  0.7486            & 0.4230 \\
~Uni3R~\cite{uni3r}           & 40.80             & 56.80             & 22.87             & 0.7760             & 0.3080             & 36.81             & 57.35              & 19.02             & 0.6817             & 0.4438 \\
~FLEG (Ours)                  & \cellbest 46.70  & \cellbest 73.04    & \cellbest 26.33   & \cellbest 0.8468   & \cellbest 0.1739   & \cellbest 48.33   & \cellbest 74.12    & \cellbest 23.62   & \cellbest 0.7920   & \cellbest 0.3219 \\

\cmidrule(r){1-6} \cmidrule(l){7-11}

\multicolumn{1}{c}{} & \multicolumn{5}{c}{16 views} & \multicolumn{5}{c}{16 views} \\
\cmidrule(r){1-6} \cmidrule(l){7-11}
~LSeg~\cite{lseg}            & \cellsecond 44.68 & 57.48             & -                  & -                  & -                  & \cellsecond 40.34 & \cellsecond 59.06 & -                  & -                  & -                  \\
\cite{sam2}+CLIP~\cite{clip} & 36.19             & \cellsecond 59.23 & -                  & -                  & -                  & 32.08             & 54.34             & -                  & -                  & -                  \\
~AnySplat~\cite{anysplat}    & -                 & -                 & \cellsecond 24.24  & \cellsecond 0.8093 & \cellsecond 0.2301 & -                 & -                 & \cellsecond 21.80  & \cellsecond 0.7551 & \cellsecond 0.3574 \\
~LSM~\cite{lsm}              & 39.66             & 57.18             & 18.28              & 0.7170             & 0.4279             & 36.85             & 57.58             & 19.82              & 0.7284             & 0.4405             \\
~Uni3R~\cite{uni3r}          & 40.29             & 55.49             & 22.08              & 0.7605             & 0.3232             & 35.53             & 56.31             & 18.69              & 0.6804             & 0.4483             \\
~FLEG (Ours)                 & \cellbest 46.09   & \cellbest 73.15   & \cellbest 24.67    & \cellbest 0.8181   & \cellbest 0.2079   & \cellbest 47.27   & \cellbest 73.65   & \cellbest 22.34    & \cellbest 0.7697   & \cellbest 0.3507   \\
\bottomrule
\end{tabularx}
\label{tab: method_comparison}
\end{table}

\begin{table}[t!]
\centering
\caption{Comparison under dense 32 input view settings on ScanNet++ and ScanNet. The \colorbox{red!23}{best} and \colorbox{orange!26}{runner-up} results are highlighted in red and orange.}
\resizebox{\columnwidth}{!}{
\scriptsize
\begin{tabularx}{0.99\textwidth}{l *{5}{Y} *{5}{Y}}
\toprule

& \multicolumn{5}{c}{ScanNet++~\cite{scannet++}} & \multicolumn{5}{c}{ScanNet~\cite{scannet}} \\
\cmidrule(r){2-6} \cmidrule(l){7-11}

& mIoU$\uparrow$ & Acc.$\uparrow$ & PSNR$\uparrow$ & SSIM$\uparrow$ & LPIPS$\downarrow$
& mIoU$\uparrow$ & Acc.$\uparrow$ & PSNR$\uparrow$ & SSIM$\uparrow$ & LPIPS$\downarrow$ \\
\cmidrule(r){1-6} \cmidrule(l){7-11}

~AnySplat\cite{anysplat}      & -                 & -                 & \cellsecond 24.08  & \cellsecond 0.8095 & \cellsecond 0.2473 & -                 & -                 & 20.91              & 0.7618              &  0.3899 \\ 
~LangSplat\cite{langsplat}    & \cellsecond 25.81 & \cellsecond 59.40 & 24.02              & 0.7749             & 0.3029             & \cellsecond 41.24 & \cellsecond 61.67 & \cellsecond 21.56  & \cellsecond 0.7974            & 0.3906  \\ 
~LangScene-X\cite{langscenex} & 21.08             & 47.61             & 23.13              & 0.7824             & 0.3435             & 27.39             & 53.33             &  20.77             & \cellbest 0.8412    & \cellsecond 0.3832 \\ 
~FLEG (Ours)                  & \cellbest 46.61   & \cellbest 67.18   & \cellbest 25.14    & \cellbest 0.8235   & \cellbest 0.2073   & \cellbest 60.97   & \cellbest 91.67   & \cellbest 23.30    & 0.7898              & \cellbest 0.3678  \\
\bottomrule
\end{tabularx}
}
\label{tab: comparison_per}
\end{table}

\begin{figure}[t!]
    \centering
    \includegraphics[width=1.0\textwidth]{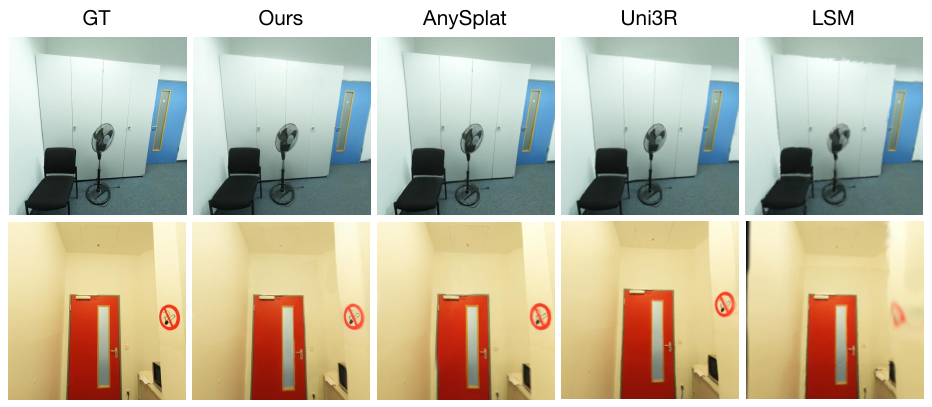}
    \caption{Qualitative comparisons on novel view synthesis.}
    \label{fig: nvs}
\end{figure}

\begin{figure}[t!]
    \centering
    \includegraphics[width=1.0\textwidth]{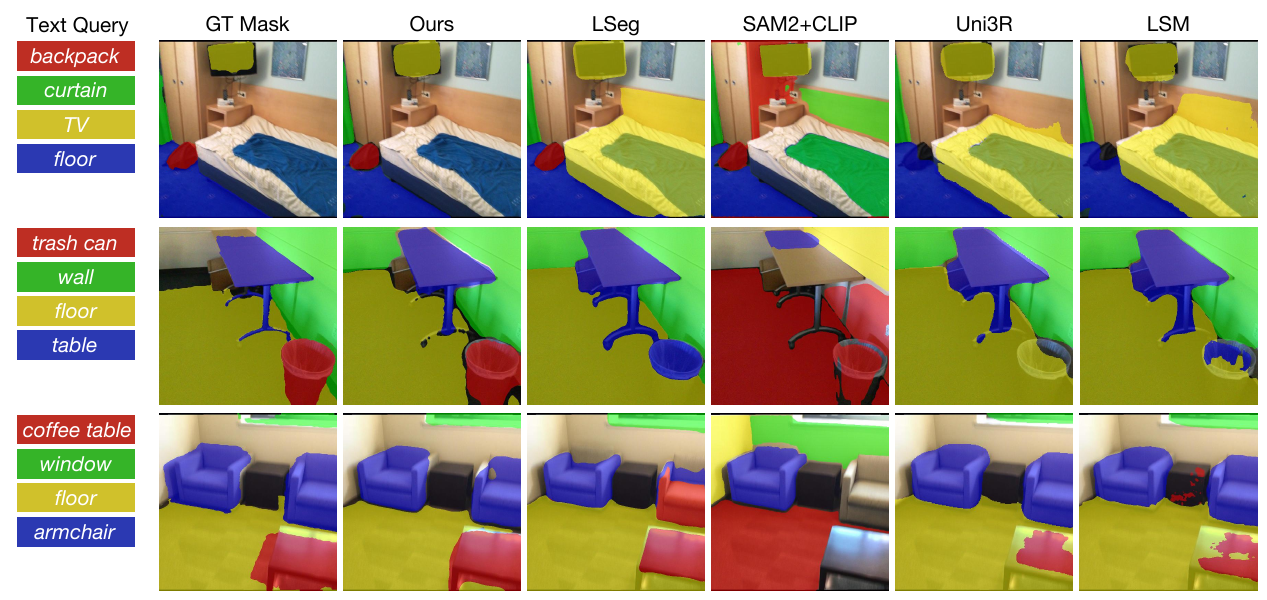}
    \caption{Qualitative comparison of open-vocabulary query segmentation.}
    \label{fig: sem}
\end{figure}

\subsection{Quantitative Comparisons}

We compare FLEG with existing feed-forward baselines on novel-view synthesis and open-vocabulary segmentation. As shown in Tab.~\ref{tab: method_comparison}, FLEG consistently outperforms other feed-forward language-embedded methods across 2–16 input views in both reconstruction and segmentation. It also surpasses commonly used 2D models employed for distillation. We also compare FLEG with AnySplat and per-scene optimized methods under dense 32 input views in Tab.~\ref{tab: comparison_per}, and our method demonstrates an overall advantage.\par

\subsection{Qualitative Comparisons}

We perform a qualitative comparison to visualize both reconstruction quality and open-vocabulary query accuracy. In Fig.~\ref{fig: nvs}, the baseline methods exhibit noticeable artifacts, whereas our method produces high-quality novel-view synthesis. In Fig.~\ref{fig: sem}, our method yields more accurate language-guided query results than baselines even with much sparser language embeddings. \par

\subsection{Ablation Studies}

We conduct ablation studies on the ScanNet++~\cite{scannet++} dataset in \cref{tab: ablation}. To verify semantic sparsity, we ablate the proposed DGLE module and conduct two per-Gaussian language-embedding variants (lines 2–3), one predicting voxelized Gaussians and the other predicting pixel-aligned Gaussians. The denser language embeddings bring only marginal improvements while incurring substantial storage overhead and slightly degrading reconstruction quality, which supports the semantic-sparsity hypothesis and validates our compact semantic representation. We also validate the dual-branch distillation framework. In the semantic distillation branch, naively aligning CLIP features without the proposed instance-guided contrastive learning (line 1 vs. line 4) produces a large drop in mIoU and Accuracy, demonstrating the necessity of distinguishing instance embeddings during training. In the geometric distillation branch, the introduced novel view distillation substantially improves both semantic representation and reconstruction quality (line 1 vs. line 5). Finally, removing geometric distillation entirely (line 6) leads to highly unstable and ineffective training. These results highlight the effectiveness of the dual-branch distillation framework. \par

\par

\begin{table}[t!]
\centering
\caption{Ablation studies on ScanNet++. Results are averaged over 2, 4, 8, and 16 input views.}
\resizebox{\columnwidth}{!}{
\scriptsize
\begin{tabularx}{0.99\textwidth}{ll *{7}{Y}}
\toprule

& & Emb. Num$\downarrow$ & Storage (MB)$\downarrow$ & mIoU$\uparrow$ & Acc.$\uparrow$ & PSNR$\uparrow$ & SSIM$\uparrow$ & LPIPS$\downarrow$ \\
\midrule
1. & Ours               & 73.89K & 18.04 & 46.87 & 72.96 & 26.66 & 0.8499 & 0.1707 \\
\midrule
2. & w/o sparse (voxel) & 1.625M  & 396.7 & 47.83 & 73.14 & 25.51 & 0.8303 & 0.1853 \\
3. & w/o sparse (pixel) & 2.012M  & 491.3 & 46.69 & 73.71 & 26.13 & 0.8415 & 0.1647 \\
\midrule
4. & w/o contrast       & 77.62K & 18.95 & 40.22 & 68.35 & 26.17 & 0.8365 & 0.1765 \\
5. & w/o novel distill  & 82.17K & 20.06 & 46.49 & 72.44 & 24.77 & 0.8165 & 0.1945 \\
6. & w/o geo. distill   & 1.001M  & 244.3 & 18.56 & 42.12 & 14.38 & 0.5966 & 0.5536 \\
\bottomrule
\end{tabularx}
}
\label{tab: ablation}
\end{table}

\section{Conclusion}

In this paper, we propose FLEG, a feed-forward network to reconstruct language-embedded Gaussians from uncalibrated and unposed multi-view images. To train the network without 3D annotations, we propose a geometric-semantic dual-branch distillation framework. We employ VGGT for supervision and propose a novel-view-based distillation strategy that mitigates overfitting to input views. For semantics, we embed CLIP features into Gaussians via semantic distillation with instance-guided contrastive learning. Importantly, we find that semantic signals are substantially sparser than geometric and appearance ones, indicating that the prevailing paradigm of attaching a language embedding to every Gaussian is redundant and incurs substantial storage overhead. Building on this insight, we propose a decoupled Gaussian language-embedding strategy that derives a separate, compact semantic representation from the standard Gaussians. Compared to dense, pixel-aligned per-Gaussian embeddings, our compact semantic representation uses less than 5\% of the language-embedding parameters while maintaining comparable semantic fidelity. This work focuses on semantic reconstruction from arbitrary viewpoints under a feed-forward paradigm and emphasizes the sparsity of semantics. It has potential future applications in embodied scenarios such as robot navigation, facilitating the perception and understanding of the 3D world via joint 3D representations. \par


%
%
\bibliographystyle{splncs04}
\bibliography{main}

\clearpage

\appendix

\section{Additional  Implementation Details}

In our training objectives, $L_{\mathrm{distill}}$ denotes the geometric distillation loss, which includes an MSE loss for depth maps and a Huber loss for camera parameters. Similar to AnySplat~\cite{anysplat}, we also introduce a depth consistency loss that consists of an MSE loss between depth maps predicted from the depth head and rendered depths. In contrast to AnySplat, we apply the geometric distillation to both the standard Gaussians and the semantic Gaussians. Notably, all geometric distillation terms are computed on target views, which are selected using our proposed novel-view-based distillation scheme to enable supervision from novel views. \par

\section{Additional Experiment Details}
In the main paper, we train and evaluate our model on the ScanNet++~\cite{scannet++} and ScanNet~\cite{scannet} datasets following their official train/validation splits. For comparison with feed-forward methods (Tab. 1 in the main paper), we evaluate on the full validation datasets. For comparison under the 32-input-view setting (Tab. 2 in the main paper), which includes per-scene optimized methods, we evaluate on a subset of scenes sampled from the validation set, since these optimization-based methods require tens of minutes to reconstruct each scene. In the supplementary material, the evaluations on the ScanNet++ dataset are performed on the full validation set. In both the main paper and the supplementary material, storage overhead is calculated assuming a default data type of float32.\par

\section{Analysis of Semantic Sparsity}

\begin{figure}[t!]
    \centering
    \includegraphics[width=0.9\textwidth]{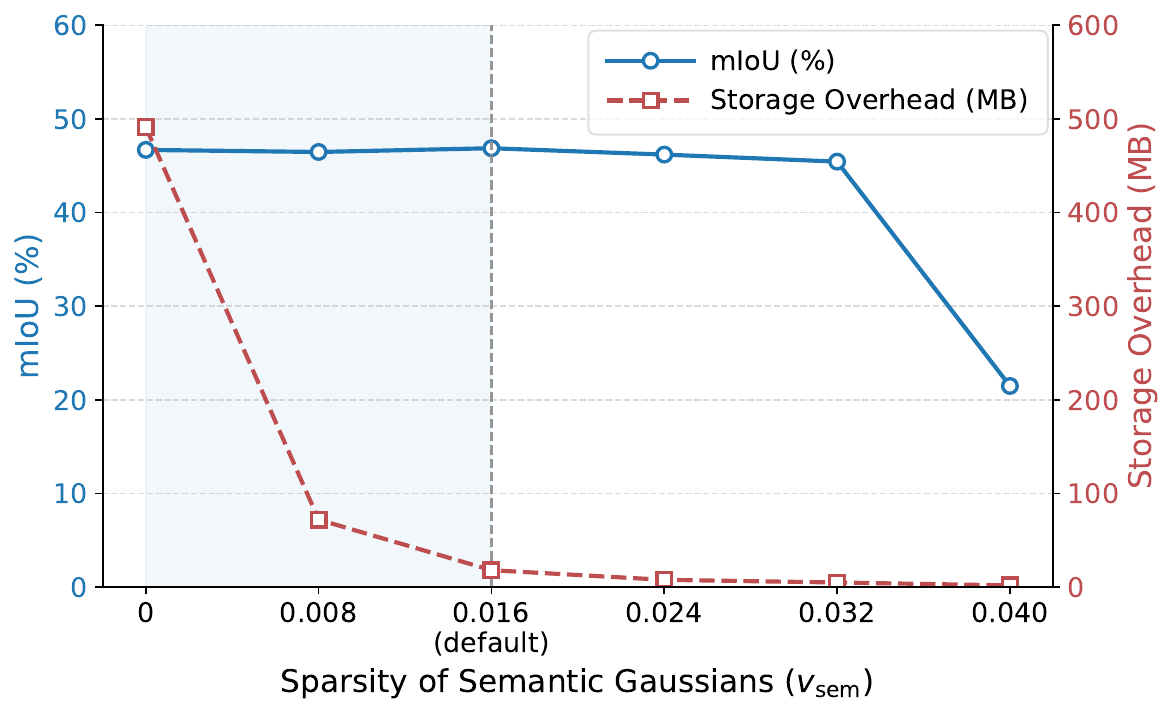}
    \caption{Impact of semantic sparsity on mIoU and storage overhead. Within the light-blue region, increasing sparsity significantly reduces the number of embeddings and corresponding storage overhead, while preserving semantic representation capability.}
    \label{fig: sparsity}
\end{figure}

A key aspect of our method is the exploitation of the sparsity of the semantic representation by introducing semantic Gaussians as a compact semantic representation. This approach avoids the need for per-Gaussian language embeddings, thereby reducing the total number of embeddings and significantly reducing storage overhead for reconstructed assets. In our method, the sparsity of semantic representation is determined by the voxel size of semantic Gaussians $v_{\mathrm{sem}}$. To systematically investigate the relationship between semantic sparsity and semantic fidelity, we conduct an additional experiment analyzing the effect of $v_{\mathrm{sem}}$ on model performance, as illustrated in Fig.~\ref{fig: sparsity}. Smaller values of $v_{\mathrm{sem}}$ correspond to denser semantic Gaussians and lower sparsity, whereas larger values of $v_{\mathrm{sem}}$ yield sparser Gaussians and higher sparsity. Increased sparsity reduces the number of required language embeddings and consequently reduces storage demands. \par

Within the light blue region of Fig.~\ref{fig: sparsity}, as $v_{\mathrm{sem}}$ increases, the mIoU remains almost unchanged, indicating that in this range, higher semantic sparsity does not lead to significant degradation of semantic fidelity. Moreover, as the semantic Gaussians become sparser, the number of language embeddings decreases substantially, resulting in a significant reduction in storage overhead. When $v_{\mathrm{sem}}$ exceeds $0.016$, mIoU begins to slightly decrease. As $v_{\mathrm{sem}}$ increases to $0.040$, although storage overhead continues to decrease, excessive sparsity hinders effective semantic representation and leads to substantial performance deterioration. \par

These observations suggest that, within a specific range, semantic sparsity can reduce the number of language embeddings and storage overhead without degrading semantic representation capability. In our experiments, we adopt $0.016$ as the default setting of $v_{\mathrm{sem}}$, which achieves significant sparsity and storage savings while preserving semantic representation capability. \par

\section{Comparison of Camera Pose Estimation}

\begin{table}[t!]
\centering
\caption{Comparison of camera pose estimation on ScanNet++ validation set. The \colorbox{red!23}{best} and \colorbox{orange!26}{runner-up} results are highlighted in red and orange.}
\footnotesize
\begin{tabularx}{0.6\textwidth}{l *{2}{Y} }
\toprule

& AUC@15$\uparrow$ & AUC@30$\uparrow$ \\
\cmidrule{1-3}  

\multicolumn{1}{c}{} & \multicolumn{2}{c}{2 views} \\
\cmidrule{1-3}  
~VGGT~\cite{vggt}                   & \cellbest 0.2733   & \cellsecond 0.4367            \\
~AnySplat~\cite{anysplat}            & 0.1653             & 0.3600                        \\ 
~FLEG (Ours)                        & \cellsecond 0.2627 & \cellbest 0.4407              \\
\cmidrule{1-3}

\multicolumn{1}{c}{} & \multicolumn{2}{c}{4 views} \\
\cmidrule{1-3}  
~VGGT~\cite{vggt}                   & \cellbest 0.4316   & \cellsecond 0.5978            \\
~AnySplat~\cite{anysplat}            & 0.2667             & 0.4728                        \\ 
~FLEG (Ours)                        & \cellsecond 0.4224 & \cellbest 0.6270              \\
\cmidrule{1-3}

\multicolumn{1}{c}{} & \multicolumn{2}{c}{8 views} \\
\cmidrule{1-3} 
~VGGT~\cite{vggt}                   & \cellsecond 0.5263 & \cellsecond 0.6980            \\
~AnySplat~\cite{anysplat}            & 0.3405             & 0.5690                        \\ 
~FLEG (Ours)                        & \cellbest 0.5358   & \cellbest 0.7223              \\
\cmidrule{1-3}

\multicolumn{1}{c}{} & \multicolumn{2}{c}{16 views} \\
\cmidrule{1-3} 
~VGGT~\cite{vggt}                   & \cellsecond 0.6238 & \cellsecond 0.7786            \\
~AnySplat~\cite{anysplat}            & 0.4663             & 0.6733                        \\ 
~FLEG (Ours)                        & \cellbest 0.6370   & \cellbest 0.7982              \\
\cmidrule{1-3}

\multicolumn{1}{c}{} & \multicolumn{2}{c}{32 views} \\
\cmidrule{1-3} 
~VGGT~\cite{vggt}                   & \cellsecond 0.6676 & \cellsecond 0.8063            \\
~AnySplat~\cite{anysplat}            & 0.5097             & 0.7045                        \\ 
~FLEG (Ours)                        & \cellbest 0.6810   & \cellbest 0.8249              \\
\bottomrule
\end{tabularx}
\label{tab: pose}
\end{table}

Our method is a pose-free, feed-forward reconstruction method that can estimate camera poses during the forward pass through the camera head. To evaluate the accuracy of the estimated camera poses, we compare our method with VGGT~\cite{vggt} and AnySplat~\cite{anysplat}, which are existing state-of-the-art pose-free, feed-forward reconstruction methods that also provide camera pose estimation capabilities. We use the standard metrics Area Under the Curve at 15 degrees (AUC@15) and Area Under the Curve at 30 degrees (AUC@30), which are based on Relative Rotation Accuracy (RRA) and Relative Translation Accuracy (RTA). RRA and RTA measure the relative angular errors in rotation and translation for each image pair. Accuracy scores are computed at varying thresholds, and the AUC is obtained as the area under the accuracy-threshold curve, taking the minimum of RRA and RTA at each threshold. \par

As shown in Tab.~\ref{tab: pose}, although both our method and AnySplat distill geometric information from VGGT, our method consistently achieves higher accuracy than AnySplat across different input views. In addition, with relatively dense input of 8, 16, and 32 views, our method achieves even better performance than VGGT. These results demonstrate the effectiveness of our distillation framework, which introduces novel-view-based distillation that leverages supervision from novel views, enhancing multi-view consistency and thereby improving the accuracy of camera pose estimation. \par

\section{Comparison of Depth Estimation}

\begin{table}[t!]
\centering
\caption{Comparison of depth estimation on ScanNet++ validation set. The \colorbox{red!23}{best} and \colorbox{orange!26}{runner-up} results are highlighted in red and orange.}
\footnotesize
\begin{tabularx}{0.95\textwidth}{l *{2}{Y} *{2}{Y}}
\toprule

& \multicolumn{2}{c}{Input Views Depth Estimation} & \multicolumn{2}{c}{Novel Views Depth Estimation} \\
\cmidrule(r){2-3} \cmidrule(l){4-5}

& AbsRel$\downarrow$ & $\delta<1.25$ $\uparrow$
& AbsRel$\downarrow$ & $\delta<1.25$ $\uparrow$ \\
\cmidrule{1-5}

\multicolumn{1}{c}{} & \multicolumn{4}{c}{2 input views} \\
\cmidrule{1-5}
~VGGT~\cite{vggt}               & \cellbest 0.0723    & \cellbest 0.9424   & - & - \\
~AnySplat~\cite{anysplat}       & \cellsecond 0.0805  & 0.9311             & \cellbest 0.0840   & \cellbest 0.9303 \\
~FLEG (Ours)                    & 0.0830              & \cellsecond 0.9317 & \cellsecond 0.0860 & \cellsecond 0.9243 \\
\cmidrule{1-5}

\multicolumn{1}{c}{} & \multicolumn{4}{c}{4 input views} \\
\cmidrule{1-5}
~VGGT~\cite{vggt}               & \cellbest 0.0686   & \cellbest 0.9405   & - & - \\
~AnySplat~\cite{anysplat}       & 0.0860             & 0.9244             & \cellsecond 0.0937 & \cellsecond 0.9141 \\
~FLEG (Ours)                    & \cellsecond 0.0773 & \cellsecond 0.9335 & \cellbest 0.0782   & \cellbest 0.9379 \\
\cmidrule{1-5}

\multicolumn{1}{c}{} & \multicolumn{4}{c}{8 input views} \\
\cmidrule{1-5}
~VGGT~\cite{vggt}               & \cellbest 0.0685   & \cellbest 0.9387   & - & - \\
~AnySplat~\cite{anysplat}       & 0.0905             & 0.9144             & \cellsecond 0.0987 & \cellsecond 0.9017 \\
~FLEG (Ours)                    & \cellsecond 0.0750 & \cellsecond 0.9333 & \cellbest 0.0781   & \cellbest 0.9270 \\
\cmidrule{1-5}

\multicolumn{1}{c}{} & \multicolumn{4}{c}{16 input views} \\
\cmidrule{1-5}
~VGGT~\cite{vggt}               & \cellbest 0.0724   & \cellbest 0.9275   & - & - \\
~AnySplat~\cite{anysplat}       & 0.0907             & 0.9128             & \cellsecond 0.1006 & \cellsecond 0.8960 \\
~FLEG (Ours)                    & \cellsecond 0.0779 & \cellsecond 0.9222 & \cellbest 0.0817   & \cellbest 0.9202 \\
\cmidrule{1-5}

\multicolumn{1}{c}{} & \multicolumn{4}{c}{32 input views} \\
\cmidrule{1-5}
~VGGT~\cite{vggt}               & \cellbest 0.0654   & \cellbest 0.9350   & - & - \\
~AnySplat~\cite{anysplat}       & 0.0862             & 0.9230             & \cellsecond 0.1132 & \cellsecond 0.9003 \\
~FLEG (Ours)                    & \cellsecond 0.0742 & \cellsecond 0.9266 & \cellbest 0.0780   & \cellbest 0.9259 \\
\bottomrule

\end{tabularx}
\label{tab: depth}
\end{table}

We further compare our method with VGGT~\cite{vggt} and AnySplat~\cite{anysplat} for the depth estimation task. For our method and AnySplat, we first reconstruct Gaussian models and render depth maps, enabling depth estimation from both input views and novel views. For VGGT, depth maps are predicted by the depth head, and depth estimation is only available for input views. We evaluate relative depth using the standard metrics Absolute Relative Error (AbsRel) and accuracy under threshold $\delta < 1.25$, which are commonly used in depth estimation. AbsRel measures the mean absolute relative difference between the predicted depth and the ground truth depth. The metric $\delta < 1.25$ measures the percentage of pixels for which the ratio between the predicted depth and the ground truth depth is within a factor of 1.25. The view selection strategy follows the same protocol as in the main paper. For each input view setting, we sample additional novel views equal to half the number of input views. All views are uniformly sampled from the video frames of the ScanNet++ validation set. \par

As shown in Tab.~\ref{tab: depth}, we report depth estimation results for both input and novel views. Since both our method and AnySplat distill geometric information from VGGT, the performance at input views is slightly lower than that of VGGT. Nevertheless, our method still performs better than AnySplat. For novel-view depth estimation, VGGT, as a point-cloud reconstruction method, does not support depth estimation at novel views. Although our method achieves slightly lower metrics than AnySplat under the extremely sparse 2-input-view setting, the gap between AnySplat’s input-view and novel-view performance increases notably as the number of input views grows, indicating potential overfitting to the input views. In contrast, our method exhibits stable performance at both input and novel views and consistently outperforms AnySplat across other input-view settings. This result may be attributed to AnySplat distilling VGGT geometry only at the input views, which may induce overfitting. In contrast, our proposed novel-view-based distillation provides supervision from novel views and effectively mitigates overfitting at the input views. \par

\section{Generalization to Unseen Dataset}

\begin{table}[t!]
\centering
\caption{Comparison on unseen Replica dataset for evaluating the generalization capability. The \colorbox{red!23}{best} and \colorbox{orange!26}{runner-up} results are highlighted in red and orange.}
\footnotesize
\begin{tabularx}{0.8\textwidth}{l *{5}{Y} }
\toprule

& mIoU$\uparrow$ & Acc.$\uparrow$ & PSNR$\uparrow$ & SSIM$\uparrow$ & LPIPS$\downarrow$ \\
\cmidrule{1-6}  

\multicolumn{1}{c}{} & \multicolumn{5}{c}{2 views} \\
\cmidrule{1-6}  
~LSeg\cite{lseg}                    & \cellsecond 28.93 & 43.52              & -                  & -                  & -  \\               
~SAM2~\cite{sam2}+CLIP~\cite{clip}  & 27.23             & \cellsecond 49.67  & -                  & -                  & -  \\                
~AnySplat\cite{anysplat}            & -                 & -                  & \cellbest 26.23    & \cellbest 0.8482   & \cellbest 0.1714    \\ 
~LSM\cite{lsm}                      & 22.56             & 40.11              & 17.38              & 0.7135             & 0.4261    \\ 
~Uni3R\cite{uni3r}                  & 24.33             & 41.15              & 23.92              & 0.8019             & 0.2877    \\ 
~FLEG (Ours)                        & \cellbest 33.95   & \cellbest 60.05    & \cellsecond 25.54  & \cellsecond 0.8253 & \cellsecond 0.1745   \\
\cmidrule{1-6}

\multicolumn{1}{c}{} & \multicolumn{5}{c}{4 views} \\
\cmidrule{1-6}  
~LSeg\cite{lseg}                    & \cellsecond 27.16 & 44.08             & -                 & -                  & -  \\           
~SAM2~\cite{sam2}+CLIP~\cite{clip}  & 25.44             & \cellsecond 51.93 & -                 & -                  & -  \\              
~AnySplat\cite{anysplat}            & -                 & -                 & \cellsecond 26.16 & \cellsecond 0.8263 & \cellsecond 0.1864    \\ 
~LSM\cite{lsm}                      & 22.67             & 39.52             & 16.86             & 0.6902             & 0.4399    \\ 
~Uni3R\cite{uni3r}                  & 24.49             & 41.06             & 22.89             & 0.7746             & 0.3073    \\ 
~FLEG (Ours)                        & \cellbest 33.14   & \cellbest 58.19   & \cellbest 26.33   & \cellbest 0.8284   & \cellbest 0.1783    \\
\cmidrule{1-6}

\multicolumn{1}{c}{} & \multicolumn{5}{c}{8 views} \\
\cmidrule{1-6} 
~LSeg\cite{lseg}                    & 28.51             & 44.17             & -                 & -                  & -  \\           
~SAM2~\cite{sam2}+CLIP~\cite{clip}  & \cellsecond 28.54 & \cellsecond 55.52 & -                 & -                  & -  \\              
~AnySplat\cite{anysplat}            & -                 & -                 & 24.17             & \cellsecond 0.7985 & \cellsecond 0.2442   \\ 
~LSM\cite{lsm}                      & 23.95             & 42.57             & 16.74             & 0.6815             & 0.4448   \\ 
~Uni3R\cite{uni3r}                  & 25.93             & 42.60             & \cellsecond 24.26 & 0.7982             & 0.2968   \\ 
~FLEG (Ours)                        & \cellbest 34.91   & \cellbest 63.20   & \cellbest 25.79   & \cellbest 0.8150   & \cellbest 0.1914 \\
\cmidrule{1-6}

\multicolumn{1}{c}{} & \multicolumn{5}{c}{16 views} \\
\cmidrule{1-6} 
~LSeg\cite{lseg}                    & \cellsecond 28.14 & 42.73             & -                 & -                        & -  \\            
~SAM2~\cite{sam2}+CLIP~\cite{clip}  & 26.72             & \cellsecond 54.81 & -                 & -                        & -  \\              
~AnySplat\cite{anysplat}            & -                 & -                 & \cellbest 24.48   & \cellbest 0.8106         & \cellsecond 0.2258   \\ 
~LSM\cite{lsm}                      & 22.50             & 40.45             & 16.35             & 0.6735                   & 0.4425   \\ 
~Uni3R\cite{uni3r}                  & 24.39             & 41.36             & 22.50             & 0.7799                   & 0.3157   \\ 
~FLEG (Ours)                        & \cellbest 33.32   & \cellbest 62.40   & \cellsecond 23.63 & \cellsecond 0.7972       & \cellbest 0.2243  \\
\bottomrule
\end{tabularx}
\label{tab: replica}
\end{table}

\begin{figure}[t!]
    \centering
    \includegraphics[width=1.0\textwidth]{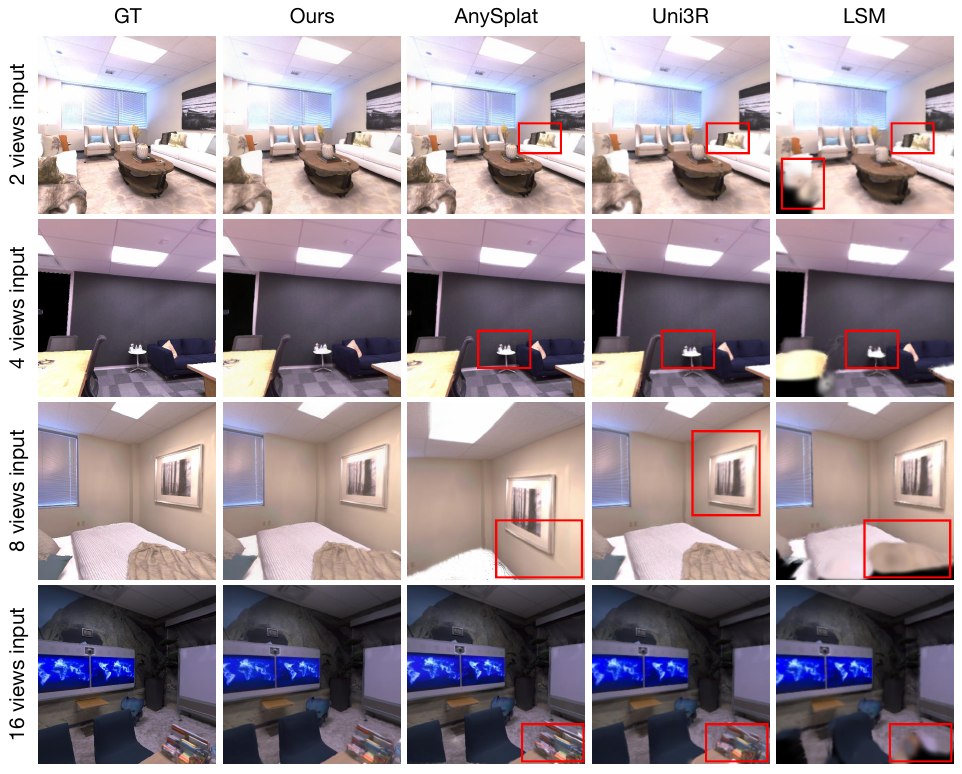}
    \caption{Qualitative comparison of novel view synthesis on the Replica dataset. Our method produces results closer to the ground truth with fewer artifacts.}
    \label{fig: nvs_replica}
\end{figure}

\begin{figure}[t!]
    \centering
    \includegraphics[width=1.0\textwidth]{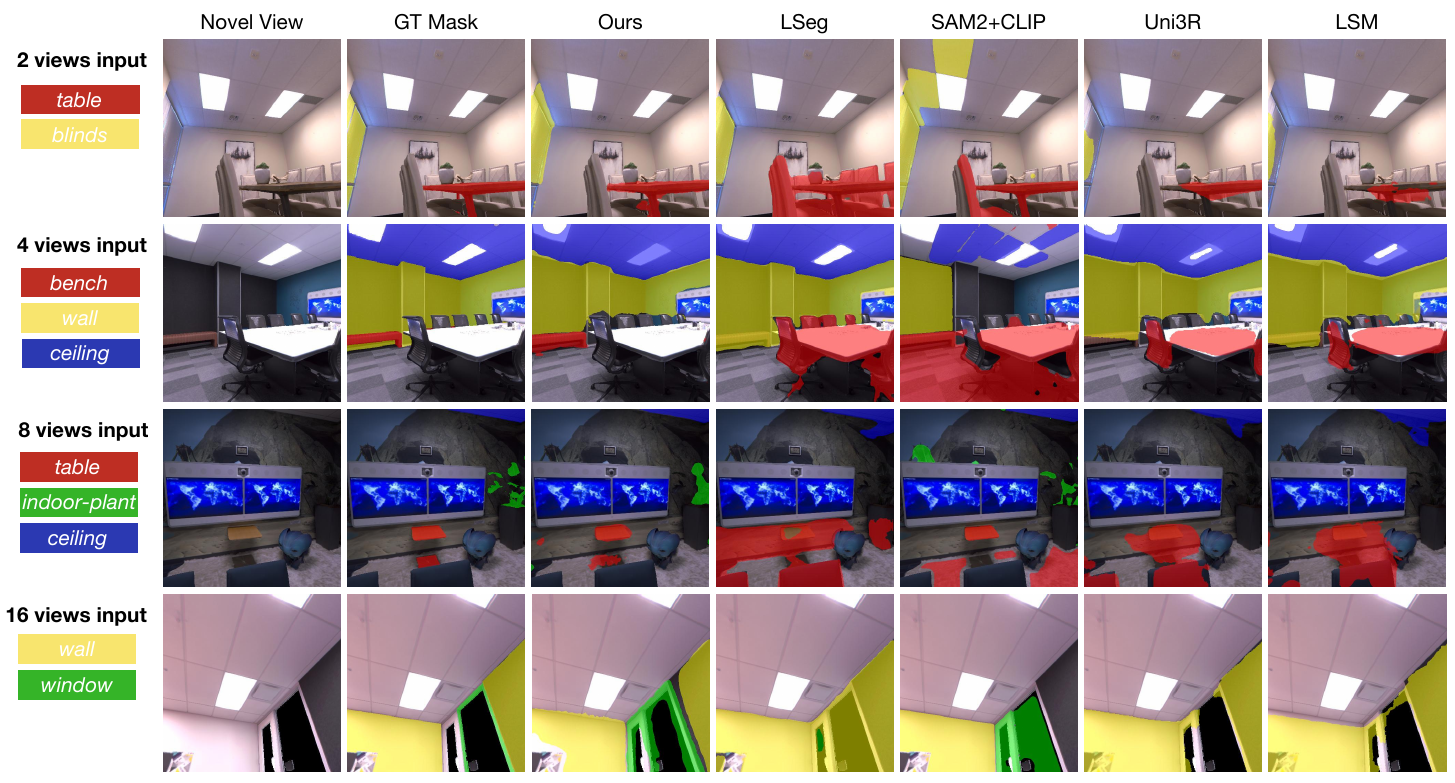}
    \caption{Qualitative comparison of open-vocabulary query segmentation on the Replica dataset. Our method produces more accurate segmentation results and better aligns with the text queries compared with baselines.}
    \label{fig: sem_replica}
\end{figure}

To evaluate the generalization capability of our method, we conduct an additional evaluation on the Replica dataset~\cite{replica}, which is not included in the training data of either our method or the baselines. The Replica dataset contains high-quality reconstructions of diverse indoor scenes, providing clean, dense geometry, high-resolution HDR textures, and annotations such as glass and mirror surfaces, planar segmentation, and semantic and instance labels. Although Replica also consists of indoor scenes, it exhibits notably different scene styles and textures from ScanNet++ and ScanNet, making it suitable for evaluating cross-dataset generalization. The comparison results are reported in Tab.~\ref{tab: replica}. \par

As shown in Tab.~\ref{tab: replica}, our model demonstrates strong generalization capability. For novel-view synthesis measured by PSNR, SSIM, and LPIPS, our method achieves a performance comparable to the existing state-of-the-art feed-forward Gaussian reconstruction method, AnySplat~\cite{anysplat}, and significantly outperforms the other baselines. Importantly, AnySplat is trained on nine large-scale mixed datasets, whereas our method is trained on only two datasets. Despite using substantially less training data, our method achieves comparable performance and additionally embeds semantic information into the Gaussians, which is not supported by AnySplat. For open-vocabulary query segmentation measured by mIoU and Accuracy, our method outperforms the baselines LSM~\cite{lsm} and Uni3R~\cite{uni3r}, which are also trained on ScanNet++ and ScanNet. Our method also outperforms CLIP~\cite{clip} with SAM2~\cite{sam2} (SAM2+CLIP), which we use for semantic distillation during our training, and LSeg~\cite{lseg}, which the baselines employ for semantic distillation. \par

In Fig.~\ref{fig: nvs_replica}, we present the qualitative comparison of novel view synthesis on the Replica dataset. Our method produces higher-quality reconstructions with finer details and generates fewer artifacts compared to the baselines. In Fig.~\ref{fig: sem_replica}, we present the qualitative comparison of open-vocabulary query segmentation on the Replica dataset. Our model produces more accurate segmentation masks across different input views and demonstrates stronger alignment with the input text queries, such as `bench' and `indoor-plant', whereas the baseline methods fail to correctly localize these objects. \par

\section{Additional Qualitative Results}

\begin{figure}[t!]
    \centering
    \includegraphics[width=0.95\textwidth]{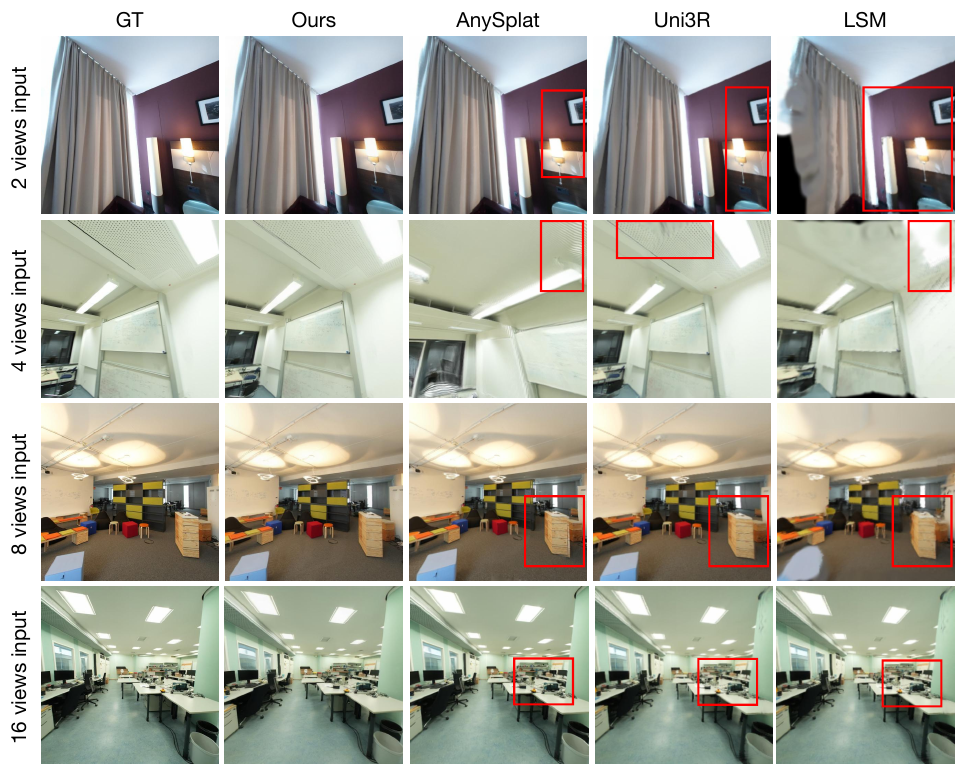}
    \caption{Additional qualitative results of novel view synthesis on the ScanNet++ validation set.}
    \label{fig: nvs_scannnetppv2}
\end{figure}

\begin{figure}[t!]
    \centering
    \includegraphics[width=0.95\textwidth]{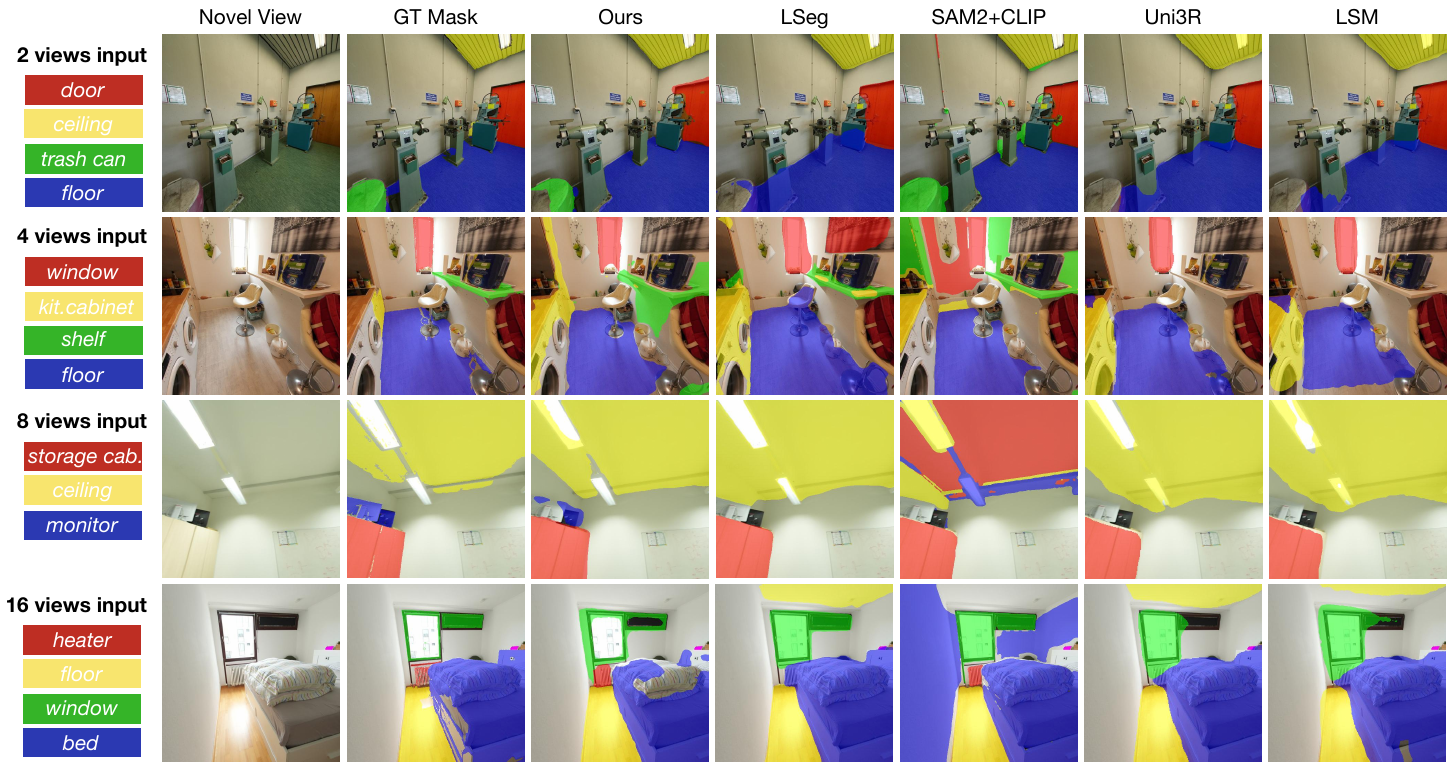}
    \caption{Additional qualitative results of open-vocabulary query segmentation on the ScanNet++ validation set.}
    \label{fig: sem_scannnetppv2}
\end{figure}

\begin{figure}[t!]
    \centering
    \includegraphics[width=0.95\textwidth]{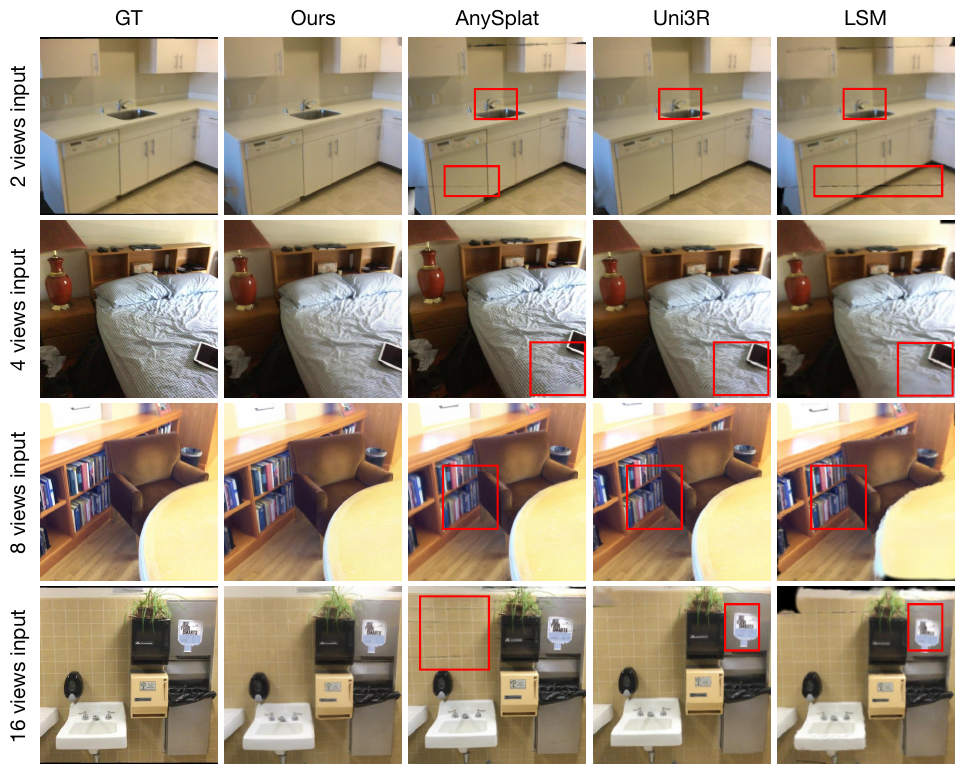}
    \caption{Additional qualitative results of novel view synthesis on the ScanNet validation set.}
    \label{fig: nvs_scannnet}
\end{figure}

\begin{figure}[t!]
    \centering
    \includegraphics[width=0.95\textwidth]{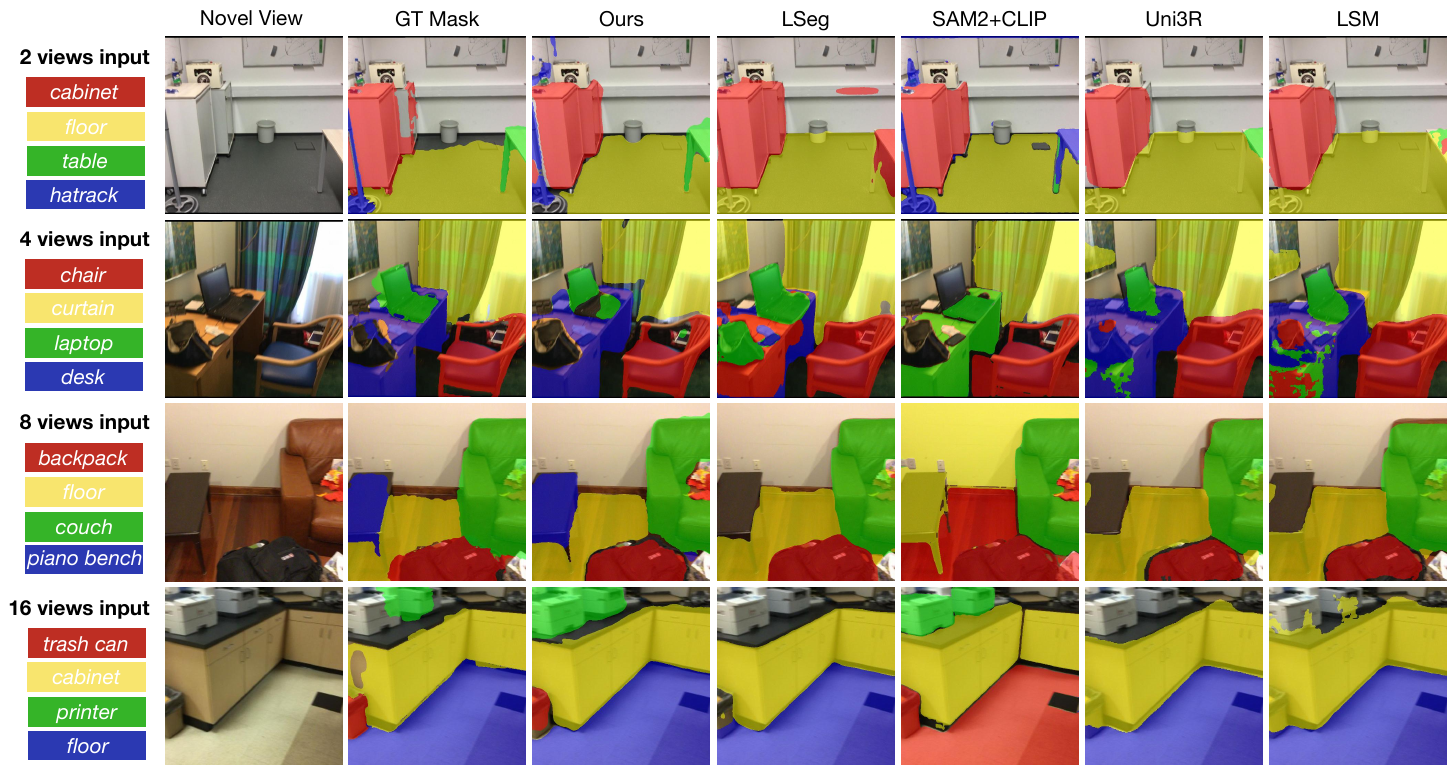}
    \caption{Additional qualitative results of open-vocabulary query segmentation on the ScanNet validation set.}
    \label{fig: sem_scannnet}
\end{figure}

\begin{figure}[t!]
    \centering
    \includegraphics[width=0.95\textwidth]{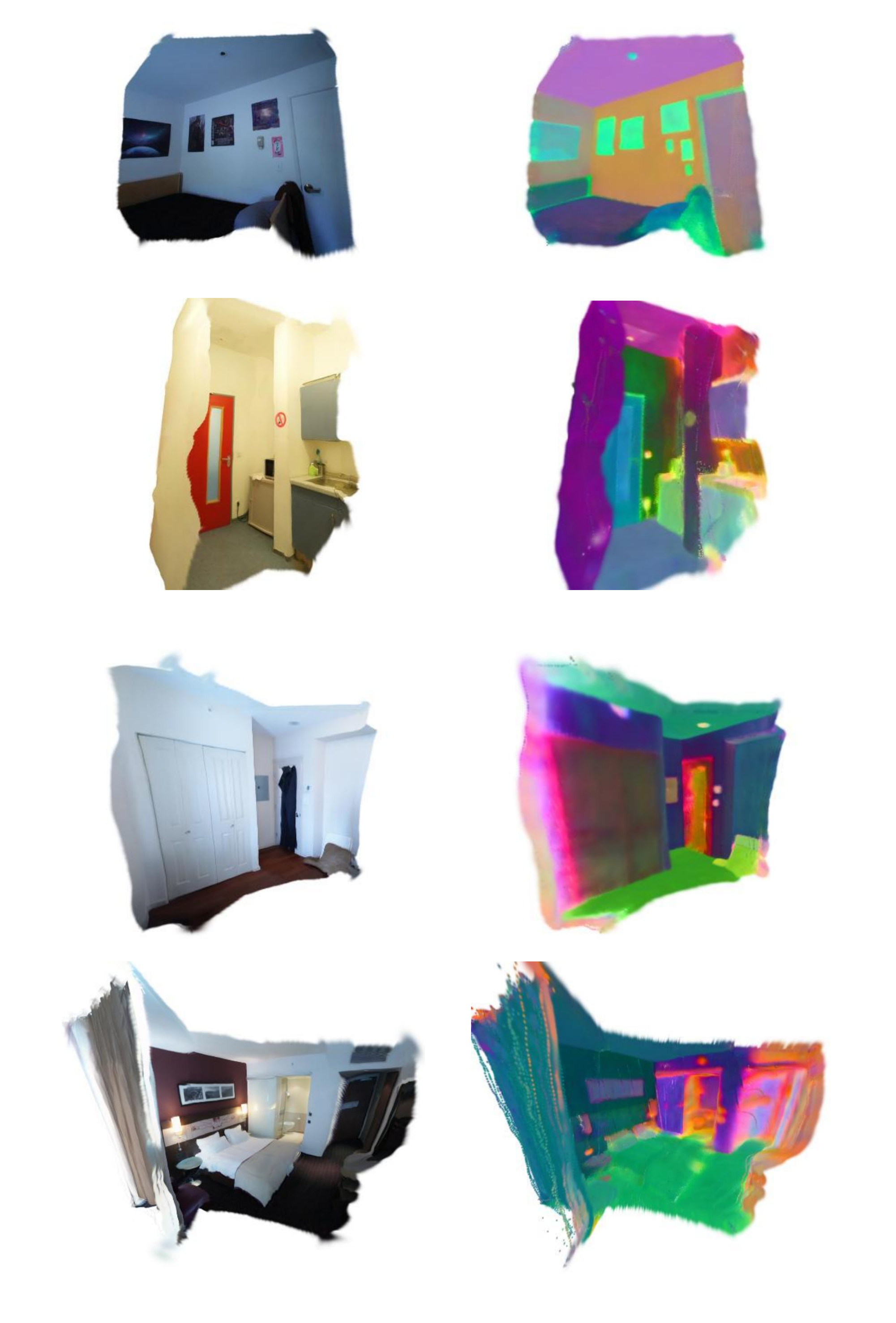}
    \caption{Visualization of 3D Gaussians and 3D Language Fields on the ScanNet++ validation set.}
    \label{fig: 3d_scannetppv2}
\end{figure}

\begin{figure}[t!]
    \centering
    \includegraphics[width=0.95\textwidth]{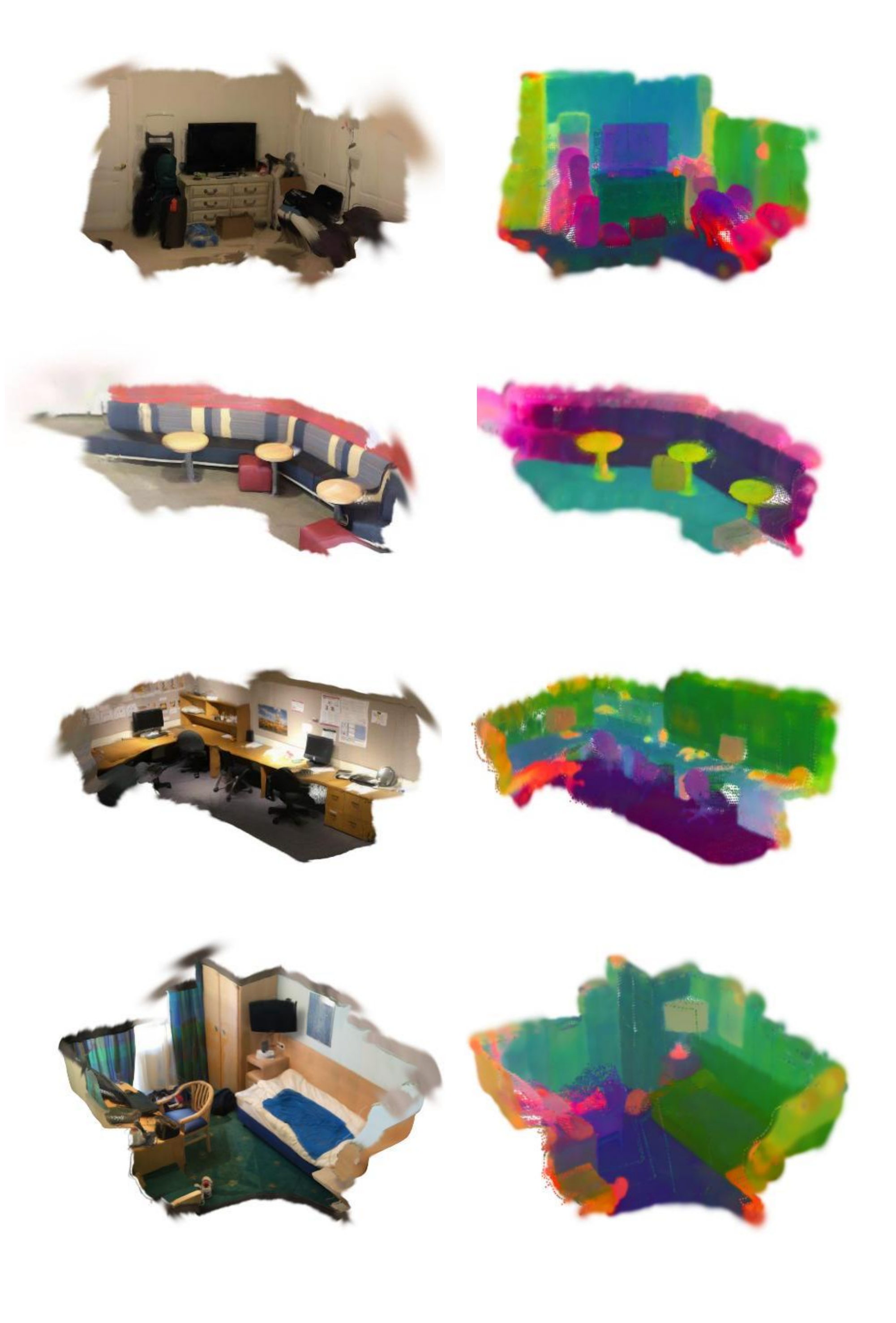}
    \caption{Visualization of 3D Gaussians and 3D Language Fields on the ScanNet validation set.}
    \label{fig: 3d_scannet}
\end{figure}

We provide more qualitative results on the ScanNet++ and ScanNet validation sets. Fig.~\ref{fig: nvs_scannnetppv2} and Fig.~\ref{fig: sem_scannnetppv2} present the novel view synthesis and open-vocabulary query segmentation results on the ScanNet++ validation set. Fig.~\ref{fig: nvs_scannnet} and Fig.~\ref{fig: sem_scannnet} present the novel view synthesis and open-vocabulary query segmentation results on the ScanNet validation set. We also provide visualizations of 3D Gaussians and 3D Language Fields on the ScanNet++ and ScanNet validation set in Fig.~\ref{fig: 3d_scannetppv2} and Fig.~\ref{fig: 3d_scannet}, respectively.\par

\section{Potential Applications}

In this paper, we propose a feed-forward language-embedded Gaussian reconstruction method that synthesizes scenes with semantic information from arbitrary views. By combining the efficiency of feed-forward reconstruction with the flexible input views, our method facilitates large-scale scene construction for applications such as semantic mapping, indoor environment modeling, and augmented/virtual reality. Our method also has potential applications in embodied systems, including robot navigation and manipulation, where accurate geometric perception needs to be integrated with high-level 3D semantic understanding. Moreover, our compact semantic representation exploits the sparsity of semantic signals, reducing storage overhead and enabling scalable, real-time deployment for both offline reconstruction and online embodied perception. \par 

\section{Limitations and Future Work}

Since our method distills geometric information from VGGT, the geometric accuracy is limited by VGGT. However, given that VGGT was trained with substantially greater data and compute resources than our method, and our priority is to implement feed-forward language embedded Gaussian Splatting, geometric distillation from VGGT is a practical choice. Besides, to allow fair comparison with the baselines, LSM and Uni3R, we train our model on the same ScanNet++ and ScanNet datasets as the baselines. As a result, although our method shows cross-dataset generalization, this generalization is mainly demonstrated on indoor scenes. Since our method does not require any 3D ground truth or semantic labels, the 2D masks used during training can be easily obtained from existing segmentation models. Therefore, one of the future works is to train on more diverse datasets that include both indoor and outdoor scenes to further improve generalization. In addition, although our method supports flexible input views and we propose a compact semantic representation, the language embeddings still introduce additional GPU memory, which makes it difficult to handle hundreds or thousands of input images. Addressing this limitation may require techniques such as model compression or streaming reconstruction, which we leave to future work. \par

\end{document}